# NodMAISI: Nodule-Oriented Medical AI for Synthetic Imaging

Fakrul Islam Tushar, PhD[1,2], Ehsan Samei, PhD[1,2], Cynthia Rudin, PhD[3], Joseph Y. Lo, PhD[1,2]

[1] Center for Virtual Imaging Trials, Carl E. Ravin Advanced Imaging Laboratories, Department of Radiology, Duke University School of Medicine, Durham, NC

[2] Dept. of Electrical & Computer Engineering, Pratt School of Engineering, Duke University, Durham

[3] Dept. of Computer Science, Duke University, Durham

**Objective:** Although medical imaging datasets are increasingly available, abnormal and annotation-intensive findings critical to lung cancer screening, particularly small pulmonary nodules remain underrepresented and inconsistently curated.

**Methods:** We introduce **NodMAISI**, an anatomically constrained, nodule-oriented CT synthesis and augmentation framework trained on a unified multi-source cohort (7,042 patients, 8,841 CTs, 14,444 nodules). The framework integrates: (i) a standardized curation and annotation pipeline linking each CT with organ masks and nodule-level annotations, (ii) a **ControlNet-conditioned rectified-flow** generator built on MAISI-v2's foundational blocks to enforce anatomy- and lesion-consistent synthesis, and (iii) **lesion-aware augmentation** that perturbs nodule masks (controlled shrinkage) while preserving surrounding anatomy to generate paired CT variants.

**Results:** Across six public test datasets, NodMAISI improved distributional fidelity relative to MAISI-v2 (real to synthetic FID range **1.18 to 2.99** vs **1.69 to 5.21**). In lesion detectability analysis using a MONAI nodule detector, NodMAISI substantially increased average sensitivity and more closely matched clinical scans (IMD-CT: **0.69** vs **0.39**; DLCS24: **0.63** vs **0.20**), with the largest gains for sub-centimeter nodules where MAISI-v2 frequently failed to reproduce the conditioned lesion. In downstream nodule-level malignancy classification trained on LUNA25 and externally evaluated on LUNA16, LNDbv4, and DLCS24, NodMAISI augmentation improved AUC by **0.07 to 0.11 at ≤20%** clinical data and by **0.12 to 0.21 at 10%**, consistently narrowing the performance gap under data scarcity.

**Conclusion:** NodMAISI enables nodule-aware CT generation and lesion-focused augmentation that reduces the distributional gap between synthetic and real screening CT while producing functionally useful training data that improves external generalization, especially when clinical data are limited.

# 1. Introduction

Medical imaging datasets are increasingly available, yet datasets enriched with well-annotated abnormalities for lung cancer screening, such as lung nodules remain underrepresented [1]. Unlike natural images, radiology images require domain-trained readers to localize and label findings, so annotation is both labor-intensive and expensive [2]. As a result, publicly released CT datasets often show heterogeneous annotation quality, incomplete lesion coverage, and inconsistent curation protocols across institutions [1]. In addition, patient privacy regulations and institutional review processes slow large-scale data sharing, so even centers with substantial screening cohorts cannot easily release datasets at scale. Researchers have addressed the shortage of high-quality labels in several ways. Early efforts mined radiology reports with rule-based NLP to create image- or study-level labels, enabling rapid classification but producing noisy, non-localizing supervision [3, 4]. Vision-language models have since been explored, but they still depend on aligned image-text pairs that are rare in medical imaging. More recent lung screening studies use hybrid pipelines in which detection models generate pseudo-labels that are then refined by experts, reducing annotation burden and limiting radiologist effort [5-7]. While virtual screening cohorts based on simulations provide valuable data, they remain constrained by finite anatomical patient models, lack of comprehensive organ textures, high computational requirements, and limited variability of physics simulation protocols (acquisition and reconstruction) [8, 9]. Moreover, controlling specific clinical variables such as nodule type, size, location, and multiplicity during cohort creation remains challenging in radiology simulator-based pipelines [8, 9].

Recent advances in diffusion models for natural image and video synthesis have prompted their extension to volumetric medical imaging, including CT [10-12]. Existing methods such as MAISI-v2 can synthesize CT volumes with controllable spatial resolution, embed anatomical labels, and impose structural constraints through organ or lesion-level masks [11]. However, their utility for lung cancer imaging is still constrained. Synthesized data may capture coarse anatomical plausibility but fail to reflect the full clinical variability. A key limitation is the generation of the most challenging, clinically relevant nodules that are small (<10 mm), low-contrast, subsolid, or juxta-pleural. Since these features are harder for diffusion models to learn and reproduce, current synthesized CT volumes may lack the nodule-level detail needed for lung cancer analysis.

In this study, we revisit CT image synthesis for lung cancer screening with a focus on four integrated components: dataset curation, generative model development, synthetic data augmentation, and downstream classification evaluation. Our main contributions are:

1) **Comprehensive lung cancer dataset:** We curated one of the largest lung nodule CT datasets by aggregating multiple open-access datasets and creating a standardized cohort. we created (i) organ-level anatomical segmentation masks via a pretrained segmentation model, (ii) pseudo-nodule detections via a detection model, and (iii) point-driven nodule segmentations. The final resource comprises 7,042 patients, 8,841 CT studies, and 14,444 nodule-level annotations, each linked to the corresponding CT volume, 3D organ segmentation, nodule segmentation mask, and 3D bounding box.

2) **Anatomically constrained, nodule-oriented CT generation:** Building on this curated cohort, we trained generative models that integrate a foundational variational autoencoder with a flow-based architecture to enable anatomically constrained, nodule-oriented CT synthesis. The models were conditioned on organ structures and nodule-level annotations that preserve global anatomy while accurately capturing nodule morphology and context.

3) **Context-aware nodule augmentation and synthesis.** Using the above generator together with nodule-mask conditioning, we augmented existing cases by modifying nodule volume while keeping the surrounding anatomy unchanged, producing longitudinal representations of nodule progression and regression for enhanced data diversity.

4) **Evaluation on lung cancer classification.** We evaluated the effect of synthesized augmentation on lung cancer classification performance across varying dataset sizes, demonstrating the utility of the generated data to improve robustness.

## 2. Methods

An overview of the complete NodMAISI pipeline used in this study is provided in **Figure 1**, highlighting dataset standardization, ControlNet-conditioned synthesis, lesion-mask augmentation, and downstream evaluation.

### 2.1 Dataset Curation and Annotation Pipeline

We constructed a unified lung nodule CT cohort by aggregating six publicly available sources: Lung Nodule Database (LNDbv4) [7], Non–Small Cell Lung Cancer Radiomics (NSCLCR) [13], Lung Image Database Consortium and Image Database Resource Initiative (LIDC-IDRI) [14], Duke Lung Cancer Screening (DLCS24) dataset [5], Integrated Multiomics (IMD-CT) [15], and LUNA25 [6]. This resulted in an aggregated dataset of 7,042 patients, 8,841 CT scans, and 14,444 nodule annotations. The original sources differed in labeling standards, lesion granularity, and annotations. Dataset composition and annotation availability are summarized in **Figure 2(a–b)**, and the overall nodule size distribution is shown in **Figure 2(c)**. Dataset-specific splits are reported in **Table S1**, with per-dataset size distributions, qualitative examples, and acquisition heterogeneity summarized in **Figures S1–S3 (Supplement)**.

This consolidation produced a single, standardized resource with mixed annotation layers. To standardize supervision across heterogeneous sources, we (i) generated organ-level masks with a pretrained segmentation model, (ii) derived pseudo-nodule candidates with a detection model, and (iii) refined nodule masks via a point-driven segmentation procedure. Each study was therefore linked to its CT volume, 3D organ segmentation, nodule mask, and 3D bounding box. **Figure S4** illustrates the end-to-end annotation workflow. To the best of our knowledge, this is one of the largest standardized lung nodule CT resources with paired body-level and nodule-level annotations.

**Lung Nodule Database (LNDbv4) [7].** The LNDbv4 dataset included 294 low-dose chest CT scans collected from a lung cancer screening population between 2016 and 2018, with patients aged 19–98 years (median age 66; 55.8% male). It contained expert-annotated nodules across various sizes and textures, from micronodules (<6 mm) to larger solid and subsolid lesions, with manual nodule segmentations linked to report metadata.

**NSCLC Radiomics dataset [13].** The NSCLCR dataset included 421 pre-treatment CT scans from non–small cell lung cancer (NSCLC) patients. From the radiation-oncologist–delineated gross tumor volumes, we derived 3D bounding boxes to localize primary lesions. The cohort had a mean age of 68.0 ± 10.1 years, 68.9% male, and a range of clinical stages and histologies including squamous cell carcinoma (36.1%), large-cell (27.1%), adenocarcinoma (12.1%), with the remaining other or not otherwise specified.

**Lung Image Database Consortium and Image Database Resource Initiative (LIDC-IDRI) [14, 16].** We used a subset of LIDC–IDRI consisting of 870 unique patient scans, each read by up to four thoracic radiologists in the original two-phase LIDC protocol. Because individual nodules often had multiple reader-provided contours, we merged (union) all available nodule masks per lesion to obtain a single segmentation, yielding 2,584 nodule annotations.

**Integrated Multiomics Dataset (IMD-CT) [15].** This Intgmultiomics dataset contained 2,032 patients, each with a chest CT and a single annotated pulmonary nodule (mean age 55.9 ± 12.3 years). Nodules were predominantly malignant (78.6%) and mostly ≥10 mm, with mean diameter 30.0 ± 20.8 mm and mean solid component 15.7 ± 13.2 mm.

**Duke Lung Cancer Screening (DLCS24) dataset [5].** The DLCS24 dataset included 1,613 low-dose chest CT scans from a clinical screening population (mean age 66 years; ~50% male). All clinically actionable nodules per Lung-RADS 2022 (typically ≥4 mm or central/airway) were annotated with 3D bounding boxes, resulting in 2,487 nodule annotations. This dataset reflected real-world screening cases with standardized, radiologist-verified nodule labels.

**LUNA25 [6].** The LUNA25 dataset consolidated low-dose CT scans from the National Lungs Screening Trials (NLST). We used the LUNA25 screening cohort consisting of 1,993 patients, 3,792 low-dose chest CT scans, and 5,899 expert-annotated nodules (≥4 mm) with malignancy verified by histopathology or long-term follow-up (≥6 years). The dataset was curated to support nodule-level malignancy risk estimation, emphasizing indeterminate nodules (5–15 mm) most relevant to clinical screening practice.

**Supplementary Section S1** provides additional dataset-specific details on cohort composition and dataset specifications, summarized in **Tables S1–S7**.

**Organ & Body Segmentations.** For each CT volume, we first obtained organ segmentation masks using the VISTA-3D [17] pretrained model and then derived a corresponding whole-body mask to complete the label space. The use of VISTA-3D ensured compatibility with MAISI [10], the baseline model that was refined for this study. VISTA-3D auto is a 3D CT segmentation foundation model that combines an automatic multi-organ branch and an interactive, point-driven branch in a shared SegResNet-style encoder-decoder [17] and was pre-trained on 11,000 CT scans to segment up to 127 classes [17]. The body mask was derived by intensity thresholding at −300 Hounsfield unit (HU) to separate patient anatomy from background air, followed by connected-component analysis to retain the largest component (the patient) and 3D hole-filling to produce a contiguous body envelope. Voxels inside the body but not assigned to any organ were given a dedicated "body" label, while voxels belonging to nodule masks were reassigned to the "nodule" label to preserve lesion priority.

**Pseudo nodule segmentation.** Starting from detection-derived bounding boxes, a point-driven 3D nodule segmentation procedure was implemented. For each candidate, a cubic volume of interest (VOI) was extracted around the original box size plus a fixed margin to include peri-nodular context. Within this VOI, intensity-based segmentation (k-means, default k=2) was applied to select the cluster with the higher mean HU as the nodule component. The resulting mask was refined using 3D morphological opening and closing to suppress noise and small gaps. Optional modules permitted mask dilation in millimeters, lobe attribution, and pleural-distance computation. Segmentation performance was evaluated using bounding boxes as the surrogate reference standard, computing Dice similarity within the VOI centered on the annotated nodule. The proposed point-driven k-means segmentation performed consistently across nodule size ranges, particularly for smaller nodules (<10 mm), where foundation models such as VISTA3D auto and VISTA3D interactive exhibited reduced accuracy. **Supplementary**

**Section S3** provides additional details of the point-driven pseudo nodule segmentation algorithm and its quantitative & qualitative evaluation.

**Segmentation Alignment and Integration.** For each CT scan, this final step combined organ masks, body contour, and refined nodule masks into a unified anatomical map (**Supplement Figure S4**). All components were spatially aligned within the same voxel grid.

## 2.2 NodMAISI Training and Inference Framework

We built the NodMAISI generative pipeline on the MAISI-v2 foundational framework [10, 11] and extended it to support anatomically constrained, nodule-oriented CT synthesis. We did not retrain the 3D variational autoencoder (VAE) and the rectified-flow backbone, which served as pretrained feature extractors and latent transport models, respectively. Instead, we trained a ControlNet conditioning branch on top of the frozen models to inject structural guidance from the body mask, nodule mask, and voxel spacing (**Figure 3**).

**Pretrained backbone (frozen).** Given a CT volume x, the pretrained VAE encodes it to a latent code to $z = E_\theta(x)$ and decodes it via $\hat{x} = D_\varphi(z)$. The VAE was originally optimized to jointly minimize a reconstruction term (to ensure the decoded CT matches the input) and a Kullback–Leibler regularization term (to keep the latent space close to a Gaussian prior). In this study, that pretrained VAE is used as is, without further finetuning. Likewise, the MAISI-v2 rectified-flow model, which predicts latent velocity fields, is kept fixed **[11].**

**ControlNet Conditioning (Trained).** To adapt the pretrained generator to lung-nodule tasks, a ControlNet module $g_\omega(.)$ was trained to map conditioning inputs $c$ into feature maps that are injected into the frozen rectified-flow U-Net [11]. The ControlNet was optimized to make the generated latent trajectory consistent with the supplied anatomy:

$$L_{\text{ctrl}} = E_t \, z[||v\psi^{\text{frozen}(z_t,t)} + g_\omega(c,t) - (z_1 - z_0)||^2]$$

where $v\psi^{\text{frozen}(z_t,t)}$ denotes the original flow and gω is the only trainable part. To improve fidelity in small lesions, a region-specific $L_{\text{reg}}$ term was added and the final loss optimized was $L_{\text{total}} = L_{\text{ctrl}} + \lambda L_{\text{reg}}$.

**Inference**. At inference time, the conditioning triplet (body, nodule, spacing) is fed through the trained ControlNet, its features are fused into the frozen MAISI-v2 flow [11], and sampling proceeds from noise $z_t \sim N(0, I)$, back to z via the rectified-flow updates. The final CT is obtained by decoding z with the pretrained VAE $\hat{x} = D_\varphi(z)$. Because only the ControlNet is trained, the method inherits the stability and coverage of MAISI-v2 while gaining fine-grained control over body and nodule structure. ControlNet was trained for 500 epochs with a batch size of 8, using a learning rate of $1\times10^{-5}$, and a class-weighted loss (weight = 100) applied to the nodule label to emphasize lesion reconstruction fidelity.

**Evaluations.** To assess the realism and clinical fidelity of the generated CT volumes, two evaluation strategies were performed: **i)** quantitative image quality analysis using the Fréchet Inception Distance (FID) [18] and **ii)** lesion detectability analysis using a pretrained nodule detection model [1, 8]. For each CT scan in a target test dataset, we produced a "**digital twin**" by conditioning the generator on that scan's anatomy and nodule masks, thus enabling direct comparisons between real and synthetic images. As a result, any differences in metrics reflected dataset-specific factors rather than mismatched conditioning.

**Image fidelity evaluation.** FID was computed between AI-generated and real CT volumes across six representative datasets: LNDbv4, NSCLCR, LIDC-IDRI, DLCS24, IMD-CT, and LUNA25. Lower FID values indicated greater similarity to the real data distribution. We reported average cross-dataset FID for real-real and real-synthetic comparisons, including both MAISI-v2 and the proposed NodMAISI generations under the same test sets.

**Lesion detectability evaluation:** To assess whether generated nodules preserved clinically relevant detectability, the MONAI nodule detection model trained on LUNA16 [1, 19] was applied to both real and synthetic scans from the IMD-CT and DLCS24 test datasets. Detection outputs were analyzed using

FROC curves, reporting sensitivity as a function of the average number of false positives per scan. To capture performance variations across lesion scales, the analysis was stratified by nodule diameter into five categories: <10 mm, 10 to <20 mm, and ≥20 mm. Each size category was evaluated independently to compare detectability trends between clinical CTs, MAISI-v2 generations, and the proposed NodMAISI generations.

## 2.3 Lesion-Aware Data Augmentation

To expand the diversity of nodule appearances within the dataset, a lesion-aware augmentation strategy was developed using the above generative model. The augmentation operates at the nodule level by modifying each nodule mask using by size scaling to simulate nodule shrinkage. 3D morphological dilation and erosion was applied iteratively until the target volume percentage was achieved, as summarized in **Algorithm 1**. The modified masks served as conditioning inputs to the ControlNet-guided generator, which synthesized corresponding CT volumes where only the lesion characteristics change while the background anatomy remained consistent. This approach can be used to create longitudinal and inter-patient variations for dataset enrichment.

To demonstrate effects of the lesion-aware augmentation, the framework was applied to the LUNA25 training and validation split. **Figure 4** illustrates the distribution shift in nodule diameters for baseline versus reduced masks, real and corresponding augmented synthetic CTs using NodMAISI.

## 2.4 Downstream Task: Lung Cancer Classification

The augmented dataset was evaluated by a downstream lung cancer classification task. LUNA25 served as the primary training cohort, while LNDbv4, LIDC-IDRI, and DLCS24 were used for external validation. **Table 1** summarizes the dataset composition, including the number of nodules, malignancy distribution, and annotation sources. Each nodule was represented as a 3D CT patch and labeled according to its malignancy status.

Since smaller nodules are underrepresented in clinical datasets, we used only the volume-reduced data from LUNA25, shrinking nodule masks by up to 50% of their original volume to emulate these smaller nodules or gradual regression in longitudinal screening. The resulting synthetic cases were combined with the original LUNA25 training data for developing downstream lung cancer classifiers.

The classification task was a binary prediction of nodule-level malignancy (cancer vs. non-cancer). It is important to note that the three test datasets differed in their labeling standards and annotation quality. LUNA16 [1, 20] and LNDbv4 [7] provided radiologist suspicion labels (RSLs) of a nodules subjectively chosen and scored by radiologists without histopathology confirmation, whereas DLCS24 [5] consists of all clinically actionable nodules with histopathology or follow-up outcomes. These datasets were intentionally selected to evaluate the usability of the synthetic data across varying label reliabilities and task difficulties. Each nodule sample was represented by a 3D patch extracted from the CT volume and labeled according to its malignancy status. A 3D ResNet50 network with randomly initialized weights was used for all experiments [1]. Input CT patches were resampled to an isotropic spacing of $0.7 \times 0.7 \times 1.25$ mm, clipped to the $[-1000, 500]$ HU range, and standardized by z-score normalization. All models used the same training schedule, data augmentations, and evaluation metrics to ensure fair comparison.

To assess the effect of synthetic data under varying data availability, the models were trained using progressively smaller fractions of the clinical dataset (100%, 50%, 20%, and 10%) under two configurations:

i. **Clinical (LUNA25):** baseline models trained only on real clinical data;
ii. **Clinical + AI-generated (100%)**: for each fraction of n% clinical data, AI-generated data equal in size to the full clinical dataset were added, e.g., "10% clinical" means 10% clinical plus 100% AI-generated.

To reduce sampling bias, each fraction experiment was repeated across all possible folds without replacement: 10 folds for the 10% subset, 5 folds for 20%, 2 folds for 50%, and 1 fold for 100%. For each fold, the model was trained from scratch, while fixing the model architecture, optimizer, and training schedule. Using the held-out test set, final performance was the nodule-level AUC averaged across folds for each data fraction.

## 3. Results

### 3.1 Evaluation of Image Fidelity and Lesion Detectability

To assess the quality of the generated CT volumes, two evaluations were conducted: image-fidelity analysis using the Fréchet Inception Distance (FID) and lesion-level detectability using a pretrained nodule-detection network.

**Image Fidelity. Table 2** presents the cross-dataset FID between real and synthetic CT volumes using the held-out test splits from six public datasets: LNDbv4, NSCLCR, LIDC-IDRI, DLCS24, IMD-CT, and LUNA25. Lower FID values denote closer alignment. **Notably, real-to-real FID also revealed substantial inter-dataset heterogeneity, with NSCLCR showing the highest cross-dataset FID against four of the other real datasets (LNDbv4, LIDC-IDRI, DLCS24, and LUNA25).** For real-to-synthetic comparisons, MAISI-v2 and NodMAISI were conditioned on each volume from that real dataset. Compared to all real datasets, the NodMAISI model consistently yielded lower (better) FID scores than the baseline MAISI-v2. Using the range of real-to-real FID values as a reference for expected clinical realism, MAISI-v2 fell out of that range for 2 of the 6 real datasets, while the NodMAISI brought it within that range for all the datasets.

**Lesion Detectability.** We applied the MONAI nodule-detection model to both real and synthetic test scans from the IMD-CT [15] and DLCS24 datasets [8] and reported FROC results (**Figure 5**). We report **average sensitivity**, defined as the mean sensitivity across the operating points of **1/8–8 false positives per scan**. Across both datasets, NodMAISI achieved higher average sensitivity and more closely matched clinical performance than MAISI-v2 (IMD-CT: NodMAISI 0.69 vs clinical 0.68 vs MAISI-v2 0.39; DLCS24: NodMAISI 0.63 vs clinical 0.57 vs MAISI-v2 0.20). In size-stratified analyses (**Figure 5**), NodMAISI improved detectability relative to MAISI-v2 across all diameter bins (<10 mm, 10–<20 mm, ≥20 mm), with the largest gap observed for **sub-centimeter nodules**. For example, on IMD-CT nodules <10 mm, MAISI-v2 detected only **2/43** lesions compared with **30/43** for NodMAISI (clinical: **43/43**), indicating that MAISI-v2 often failed to reproduce the smallest lesions. Differences in dataset size composition (**Figure S1**) likely contribute to the overall performance differences between IMD-CT and DLCS24. Detailed operating-point sensitivities and detection rates are provided in **Tables S8–S9** (**Supplement**). **Figure 6** provides qualitative examples in which MAISI-v2 **fails to synthesize the conditioned target nodule** (missing or markedly attenuated lesion in the generated CT), whereas NodMAISI reproduces the nodule at the expected location and extent, consistent with the detectability gaps observed in **Figure 5**.

## 3.2 Lesion-aware augmented Synthetics CT

We applied lesion-aware augmentation to the LUNA25 training and validation splits (Table S7, Supplement) by perturbing the **nodule/lesion mask** while keeping the **organ/anatomy mask fixed**, thereby preserving patient anatomy and modifying only lesion characteristics. Using these augmented masks as conditioning inputs, we generated corresponding synthetic CT variants with NodMAISI. As shown in **Figure 4(a)**, augmentation systematically modulated the nodule diameter distribution, shifting the size statistics relative to the original clinical cohort (clinical mean/median: 17.0/16.2 mm; augmented mean/median: 13.3/12.5 mm). **Figure 4(b)** provides qualitative examples demonstrating that the augmented synthetic CTs maintain anatomical context while reflecting the intended lesion changes (purple contours) relative to the original clinical lesions (cyan), enabling controlled diversification of lesion presentations for downstream experiments.

## 3.3 Downstream Classification Performance

We evaluated whether synthetic CT augmentation improves downstream lung cancer classification when clinical training data are limited. Models were developed on **LUNA25** and externally evaluated on **LUNA16, LNDbv4, and DLCS24**. To emulate data scarcity, we trained models using **100%, 50%, 20%, and 10%** of the available clinical training set and compared three training conditions: **clinical-only**, **clinical + MAISI-v2 synthetic CT**, and **clinical + NodMAISI synthetic CT** (Figure 7; Table 3). Across all three external test sets, the **clinical-only** baseline showed the expected degradation as the fraction of clinical training data decreased. For example, on **LUNA16**, AUC decreased from **0.86** at 100% clinical data to **0.59** at 10% (Table 3). When ample clinical training data were available (100%), synthetic augmentation produced **modest** changes overall, with the clearest benefit observed on **LNDbv4**, the smaller and more distribution-shifted external test set.

In contrast, augmentation benefits became pronounced in the **low-data regime**. At **≤20%** clinical training data, **NodMAISI consistently achieved the highest AUC** across external test sets (Figure 7; Table 3). For instance, at **10%** clinical training data, NodMAISI improved AUC to **0.80** on LUNA16 (vs **0.59** clinical-only; **0.77** MAISI-v2), to **0.78** on LNDbv4 (vs **0.62** clinical-only; **0.74** MAISI-v2), and to **0.65** on DLCS24 (vs **0.53** clinical-only; **0.63** MAISI-v2). Fold-level results with **95% confidence intervals** are reported in **Tables S10–S12** (Supplement).

Overall, these results show that **nodule-oriented synthetic augmentation** provides the greatest utility when clinical data are scarce, consistently narrowing the performance gap relative to models trained with larger clinical datasets and outperforming baseline MAISI-v2 augmentation in most low-data settings.

## 4. Discussion

This study introduced NodMAISI, a nodule-oriented generative framework developed using one of the largest curated clinical lung nodule datasets. The work made four contributions. First, a standardized multi-dataset cohort was assembled with organ and nodule annotations. Second, a lesion-aware augmentation strategy was designed to systematically vary nodule morphology while preserving surrounding anatomy, effectively expanding the dataset with clinically plausible variability. Third, the NodMAISI generator was trained on this curated dataset and evaluated across multiple clinical test sets to assess image distributional fidelity and nodule detectability, demonstrating improvements compared to the baseline. Finally, the utility of the synthetic data was validated through a downstream lung cancer classification task, where models trained with NodMAISI-augmented data achieved higher performance when real data was scarce. Together, these contributions establish an end-to-end framework for generating, evaluating, and applying anatomically constrained synthetic CT data in lung cancer AI workflows.

The lesion-detectability analysis showed that NodMAISI produced nodules that remained detectable by a CAD algorithm, whereas MAISI-v2 often failed to reproduce small lesions. Size-stratified analyses also favored NodMAISI, especially for sub-centimeter nodules. In the downstream classification experiments, adding NodMAISI generated nodules mitigated the performance degradation expected from reduced clinical training data. NodMAISI consistently outperformed both the clinical-only and MAISI-v2-augmented models, with the most pronounced benefits under severe data scarcity (≤20% of the clinical dataset). These results suggest that anatomically constrained, pathology-aware generation can effectively extend limited clinical datasets without introducing domain drift, supporting the broader goal of data-efficient AI development in medical imaging.

While promising, several limitations warrant mention. First, compared with biology- and physics-based virtual imaging trials, this approach offers a much faster and more accessible alternative for generating CT data, as it does not rely on complex multi-component frameworks such as XCAT phantoms or DukeSim scanner simulations. However, this advantage comes at the cost of reduced controllability; it cannot fully emulate variations such as scanner hardware, acquisition protocols, or reconstruction kernels. Second, although generation was conditioned on nodule masks, it lacks explicit control over nodule phenotype, such as solid, subsolid, and ground-glass types. Incorporating phenotype conditioning or style tokens could enable more targeted synthesis. Finally, the current evaluation focused on detectability and malignancy classification. Future work should assess impacts on other downstream tasks such as segmentation, radiomics feature stability, longitudinal progression modeling, and clinical workflow integration.

In conclusion, NodMAISI demonstrated that clinically grounded, nodule-aware CT synthesis can both narrow the distributional gap to real screening data and produce synthetic cases that are useful for downstream learning. The framework enabled rapid expansion of training data without requiring full imaging simulation and shows promise for data-efficient lung cancer AI.

**Acknowledgments**

This work was supported by the Center for Virtual Imaging Trials, NIH/NIBIB P41 EB028744, NIH/NIBIB R01 EB038719, and the Putman Vision Award awarded by the Department of Radiology of Duke University School of the Medicine. Data was derived from the Duke Lung Cancer Screening Program.

**Data and Code Availability**

We have publicly released all code, pretrained models, and baseline results associated with this study. These resources are available at the following repositories:

**GitHub**: https://github.com/fitushar/NoMAISI

**Hugging Face**: https://huggingface.co/ft42/NoMAISI

**LUNA16**: https://luna16.grand-challenge.org/Data/

**LUNA25**: https://luna25.grand-challenge.org/

**Algorithm 1.** Lesion Volume Adjustment for Lesion-Aware Augmentation.

---

**Volume Reduction**: Lesion Shrinking via Controlled Dilation Reversal

---

**Input:**
- Binary lesion mask M ∈
- Target shrinkage percent S% (volume to reach)
- Connectivity c (default = 1)
- Max iterations N (default = 50)

**Procedure:**
1. Compute $V_{orig} = \sum M[x, y, z]$
2. Initialize: $M_{current} \leftarrow M$, $M_{best} \leftarrow M$, $\delta_{best} \leftarrow |100 - S|$
3. Repeat up to N times:
   a. Apply binary dilation to $M_{current}$
   b. Compute current volume $V_{current}$ and $P = 100 \times V_{current} / V_{orig}$
   c. Update best mask if $|P - S| < \delta_{best}$
   d. Stop if $P \geq S$ and diff > $\delta_{best}$
4. Return $M_{best}$ as shrunken mask
5. Refill removed lesion voxels using nearest lobe ID (28–32)

---

**Table 1: Summary of Datasets Used for Downstream Lung Cancer Classification Evaluation.**

| Dataset | # Nodules (positive n, %) | Annotation Source |
|---|---|---|
| **Development Dataset** | | |
| LUNA25 | 6163 (555; 9%) | Histopathology & follow-up |
| **External Test** | | |
| LUNA16 | 677 (327; 48%) | Visual Scoring |
| LNbv4 | 589 (74; 12%) | Visual Scoring |
| DLCS24 | 2487 (264; 11%) | Histopathology & follow-up |

**Table 2: Fréchet Inception Distance (FID) of real and synthetic CT distributions across datasets.**
Columns indicate the reference/test dataset used to compute FID. The **Real** block reports cross-dataset FID between real images from different datasets, and the **Synthetic** block reports FID between synthetic images generated by **MAISI-v2** or **NodMAISI** and each reference dataset. Lower FID indicates greater similarity. **Bold** values mark the **worst-performing** result for each reference dataset (highest FID).

| **FID (Avg.)** | | LNDbv4 | NSCLCR | LIDC-IDRI | DLCS24 | IMD-CT | LUNA25 |
|---|---|---|---|---|---|---|---|
| **Real** | LNDbv4 | | 5.13 | 1.49 | 1.05 | 2.40 | 1.98 |
| | NSCLCR | **5.13** | | **3.12** | **3.66** | 1.56 | **2.65** |
| | LIDC-IDRI | 1.49 | 3.12 | | 0.79 | 1.44 | 0.75 |
| | DLCS24 | 1.05 | 3.66 | 0.79 | | 1.56 | 1.0 |
| | IMD-CT | 2.40 | 1.56 | 1.44 | 1.56 | | 1.57 |
| | LUNA25 | 1.98 | 2.65 | 0.75 | 1.0 | 1.57 | |
| | | | | | | | |
| **Synthetic** | MAISI-v2 | 3.15 | **5.21** | 2.70 | 2.32 | **2.82** | 1.69 |
| | NodMAISI | 2.99 | 3.05 | 2.31 | 2.27 | 2.62 | 1.18 |

**Table 3.** AUC performance of Clinical, Clinical + MAISI-v2, and Clinical + NodMAISI models tested on three external datasets (LUNA16, LNDbv4, and DLCS24). Models were trained with varying fractions of the clinical dataset (100%, 50%, 20%, and 10%). **Bold values indicate the highest AUC within each external test dataset for a given percentage of clinical training data.**

| Models | Percentage of Clinical training Data | | | |
|---|---|---|---|---|
| | 100% | 50% | 20% | 10% |
| External Test Data: **LUNA16** | | | | |
| Clinical (LUNA25) | **0.86** | 0.81 | 0.72 | 0.59 |
| Clinical + MAISI-v2 | **0.86** | 0.83 | 0.79 | 0.77 |
| Clinical + NodMAISI | **0.86** | **0.86** | **0.81** | **0.80** |
| External Test Data: **LNDbv4** | | | | |
| Clinical (LUNA25) | 0.73 | 0.73 | 0.67 | 0.62 |
| Clinical + MAISI-v2 | **0.80** | 0.76 | 0.75 | 0.74 |
| Clinical + NodMAISI | 0.79 | **0.81** | **0.78** | **0.78** |
| External Test Data: **DLCS24** | | | | |
| Clinical (LUNA25) | 0.71 | 0.67 | 0.60 | 0.53 |
| Clinical + MAISI-v2 | 0.70 | 0.66 | 0.63 | 0.63 |
| Clinical + NodMAISI | **0.72** | **0.68** | **0.67** | **0.65** |

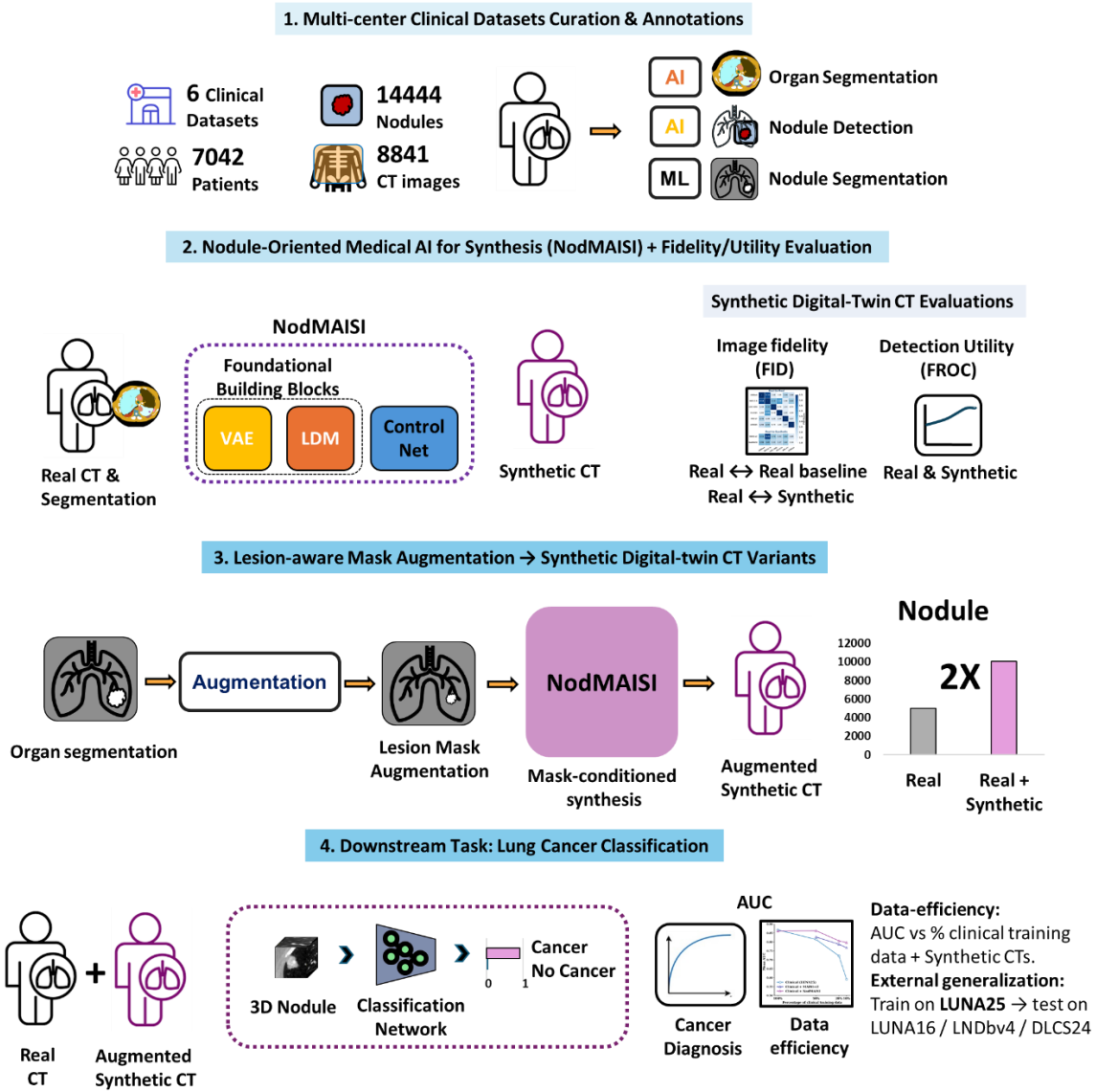

**Figure 1. NodMAISI workflow for synthetic CT generation, evaluation, and downstream use.** Multi-center clinical CT datasets are curated and annotated to obtain organ segmentation and nodule labels. NodMAISI (VAE + latent diffusion + ControlNet conditioning) synthesizes mask-conditioned CTs from real CT/masks, which are evaluated for **image fidelity** (FID; real↔real vs real↔synthetic) and **detection utility** (FROC). Lesion-aware **mask augmentation** perturbs the lesion mask while preserving anatomy to generate synthetic CT variants, increasing lesion diversity. Real and synthetic data are then used for the downstream lung cancer classification task, reporting AUC, data-efficiency under reduced clinical training fractions, and external generalization.

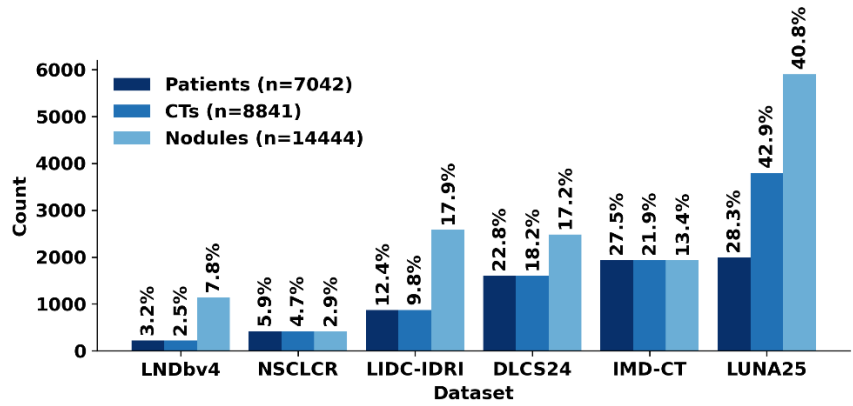

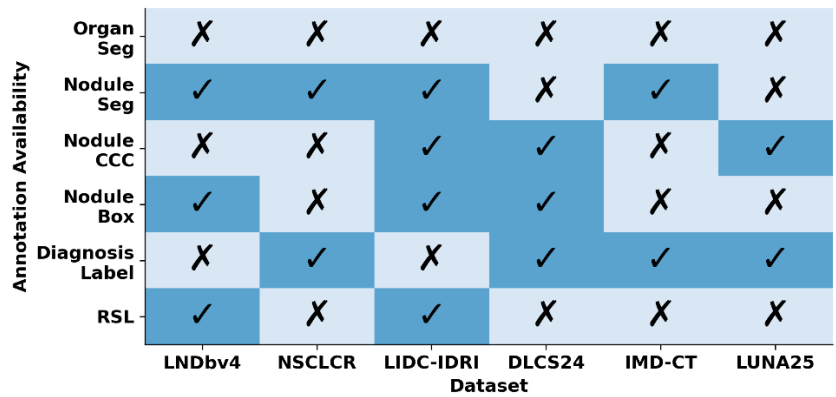

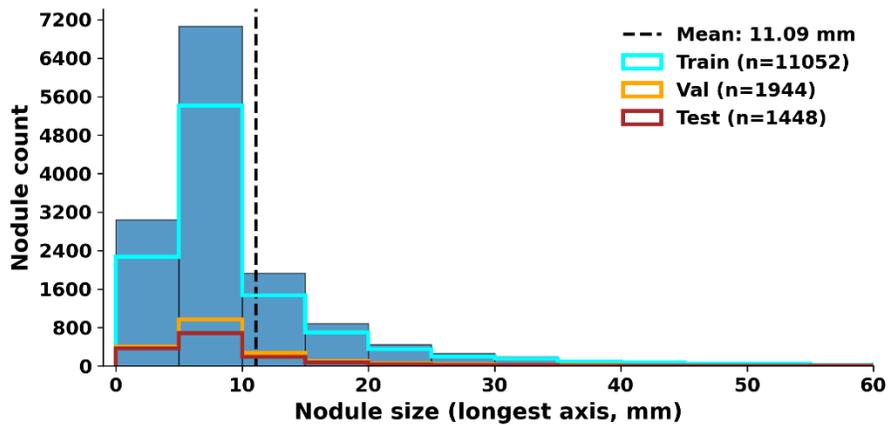

**Figure 2. Summary of the unified lung nodule CT cohort and annotation availability.** (a) Dataset composition by number of patients, CT scans, and nodules across LNDbv4, NSCLCR, LIDC-IDRI, DLCS24, IMD-CT, and LUNA25. (b) Availability of annotation layers by dataset (e.g., organ segmentation, nodule segmentation, bounding boxes, diagnosis labels). (c) Nodule size distribution (longest-axis diameter) across train/validation/test splits.

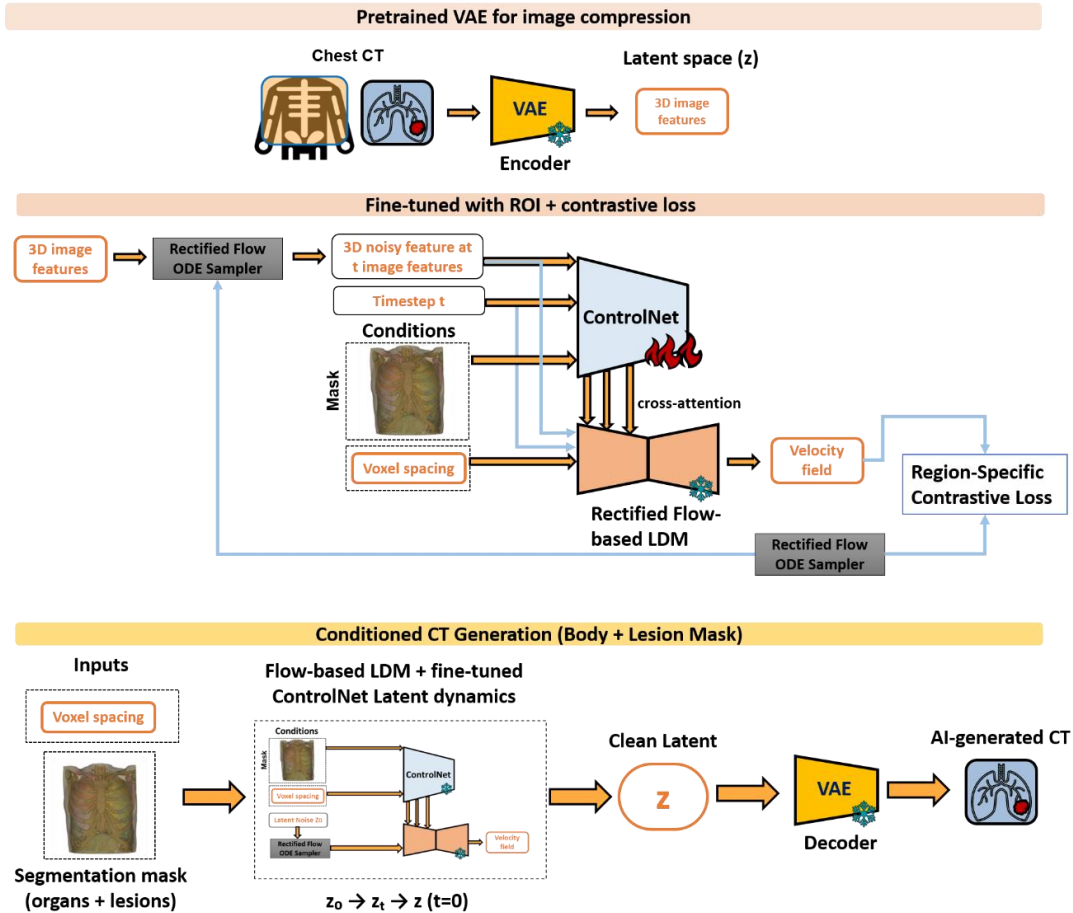

**Figure 3**: Overview of our flow-based latent diffusion model with ControlNet conditioning for AI-based CT generation. The pipeline consists of three stages: (top) Pretrained VAE for image compression, where CT images are encoded into latent features using a frozen VAE; (middle) Model fine-tuning, where a Rectified Flow ODE sampler, conditioned on segmentation masks and voxel spacing through a fine-tuned ControlNet, predicts velocity fields in latent space and is optimized with a region-specific contrastive loss emphasizing ROI sensitivity and background consistency; and (bottom) Inference, where segmentation masks and voxel spacing guide latent sampling along the ODE trajectory to obtain a clean latent representation, which is then decoded by the VAE into full-resolution AI-generated CT images conditioned by body and lesion masks

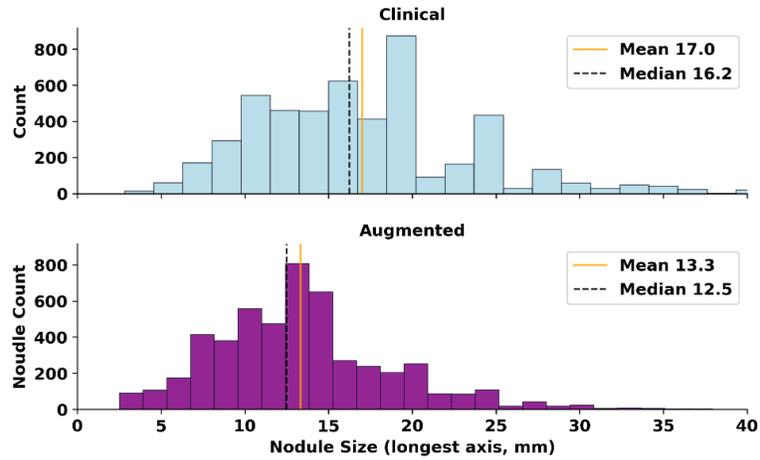

(a)

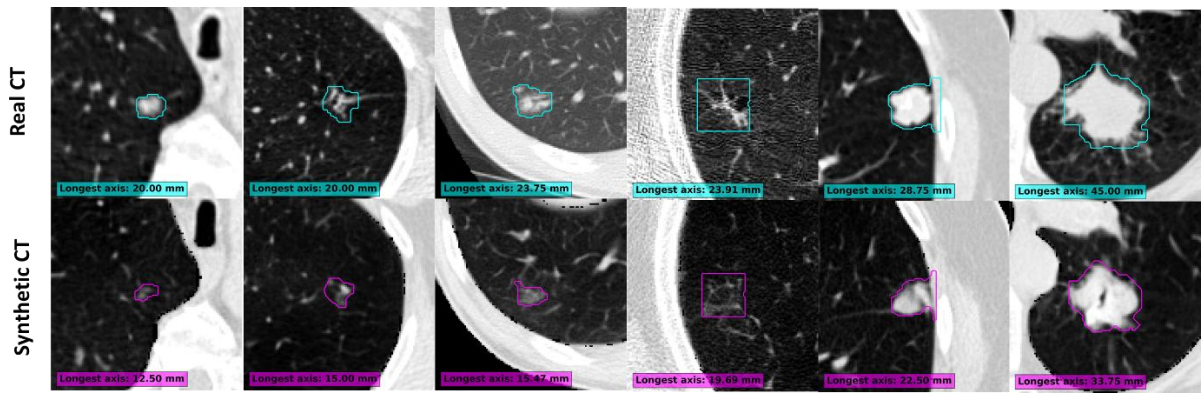

(b)

**Figure 4:** Lesion-aware nodule augmentation on LUNA25 (pseudo-segmentation baseline). Histogram plot comparing baseline versus augmented nodule diameters. Augmented nodules include reduced (purple), demonstrating systematic size modulation relative to baseline (cyan). Mean and median diameters are indicated, showing a leftward shift with shrinkage and a rightward shift with expansion relative to the baseline distribution. (b) Shown some augmented synthetic generated CT with nodule (contour in purple) relative to the original clinical CT (cyan).

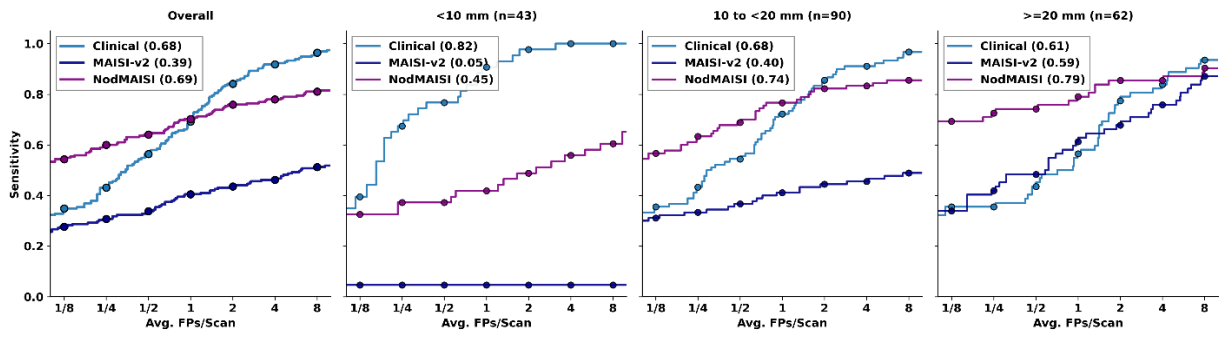

(a)

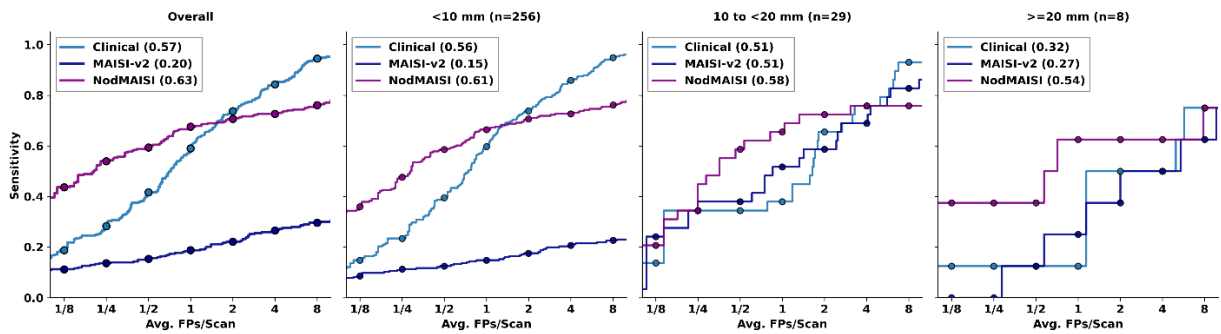

(b)

**Figure 5.** Lesion detectability evaluation on synthetic and clinical CTs. (First column) Freeresponse receiver operating characteristic (FROC) curves comparing detection sensitivity versus false positives per scan on the (a) IMD-CT and (b) DLCS24 test datasets. (second to forth columns) NodMAISI demonstrates higher sensitivity at comparable false-positive rates relative to MAISI. Size-stratified FROC analysis across nodule diameter ranges (<10 mm, 10 to <20 mm, ≥20 mm)

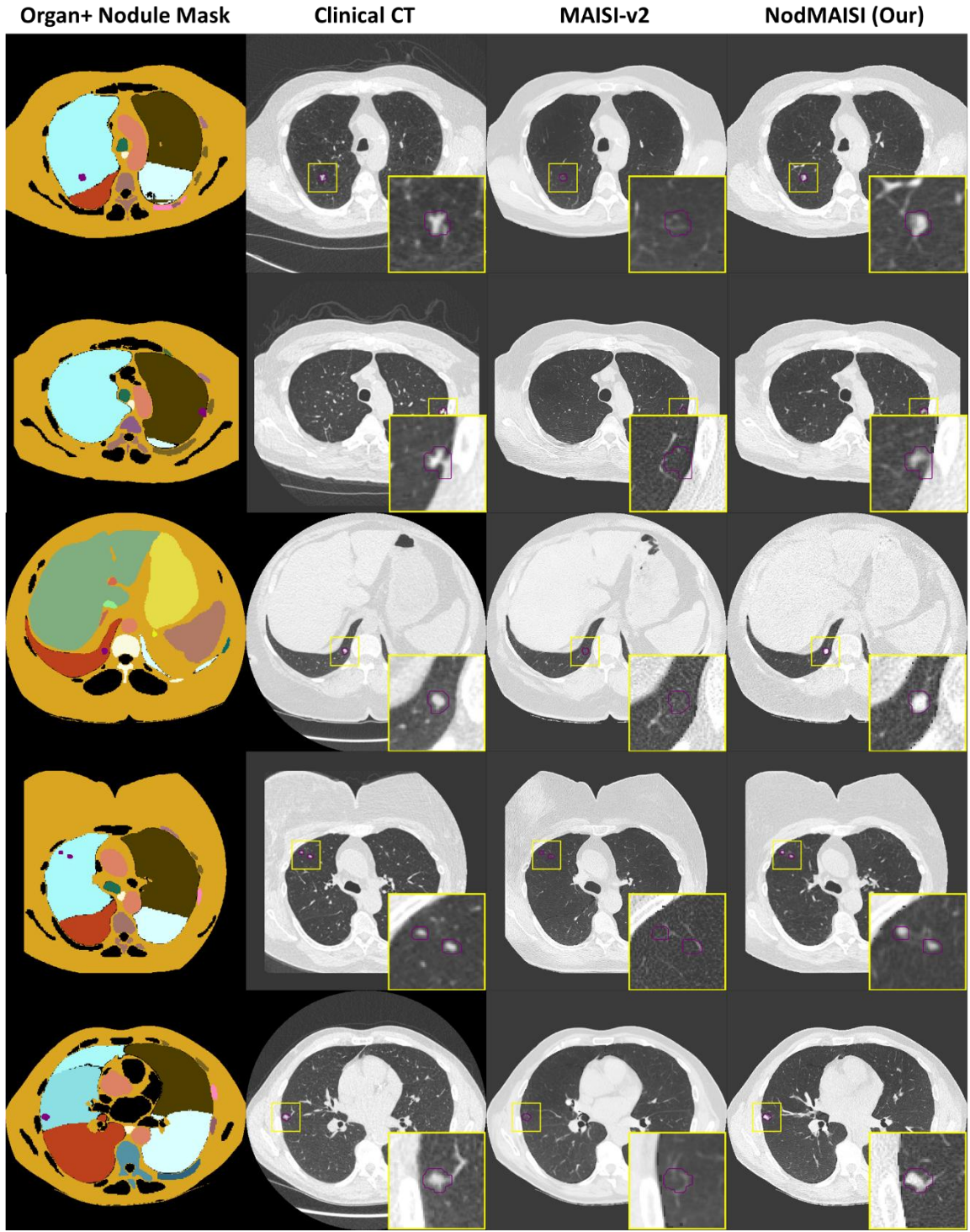

**Figure 6: Qualitative visualization of missed nodule cases during detectability analysis.**
From left to right: (left) input anatomical and lesion mask (purple) used for conditioning, (second) real clinical CT, (third) MAISI-v2 generated CT slice showing failure to reproduce the target nodule and (right) NodMAISI output depicting successful nodule generation.

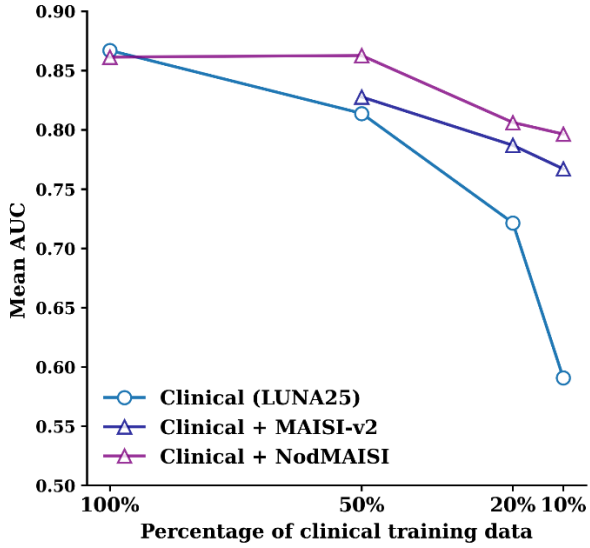

**(a) LUNA16**

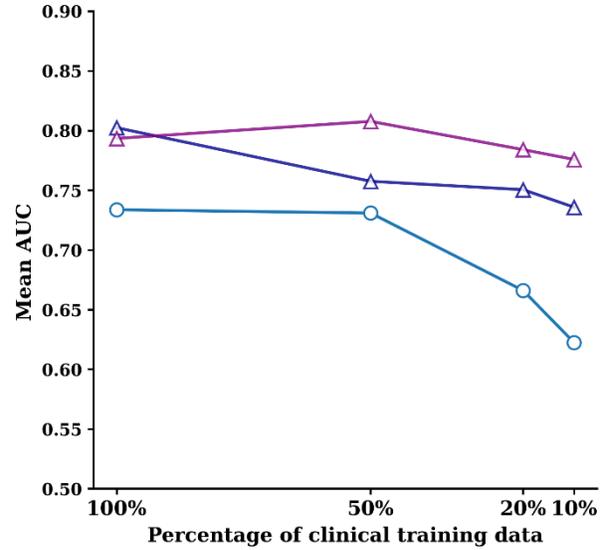

**(b) LNDbv4**

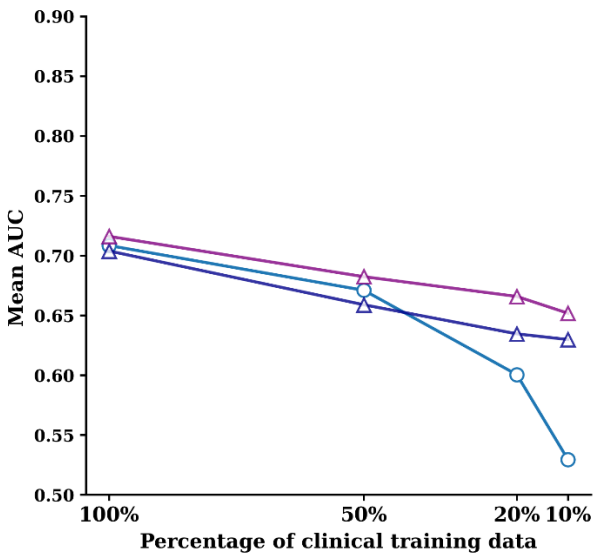

**(c) DLCS24**

**Figure 7.** Performance of lung cancer classification models trained with different proportions of clinical data and synthetic augmentation on (a) LUNA16, (b) LNDbv4, and (c) DLCS24 test sets. NodMAISI consistently improves AUC, especially under limited clinical data conditions ($\leq 20\%$).

# Supplementary Materials

# NodMAISI: Nodule-Oriented Medical AI for Synthetic Imaging

Fakrul Islam Tushar, PhD[1,2], Ehsan Samei, PhD[1,2], Cynthia Rudin, PhD[3], Joseph Y. Lo, PhD[1,2]

[1] Center for Virtual Imaging Trials, Carl E. Ravin Advanced Imaging Laboratories, Department of Radiology, Duke University School of Medicine, Durham, NC, USA
[2] Department of Electrical & Computer Engineering, Pratt School of Engineering, Duke University, Durham, NC, USA
[3] Department of Computer Science, Duke University, Durham, NC, USA

This document provides supplementary methods, additional experiments, and extended quantitative results referenced in the main manuscript, including dataset-specific cohort summaries (Tables S1–S7), acquisition heterogeneity analyses (Figures S1–S3), segmentation algorithm details and evaluation (Section S2), and fold-level external validation results with 95% confidence intervals (Tables S10–S12).

## Contents

# S1. NodMAISI: Development and Evaluation dataset:

Supplementary **Table S1** summarizes the cohort composition and data splits used in this study. For each dataset, we report the number of patients, CT scans, and annotated nodules allocated to the training, validation, and test sets. These splits form the basis for all model development and evaluation, and the totals in the last row reflect the overall scale of our benchmarking cohort.

**Table S1.** Composition of training, validation, and test splits across datasets. Number of patients, CT scans, and annotated nodules for the train/validation/test sets in each dataset. The final row summarizes the total number of patients, CT exams, and nodules used across all datasets.

| Dataset | Train/Validation/ Test dataset | | |
|---|---|---|---|
| | **Patients (#)** | **CTs (#)** | **Nodules (#)** |
| LNDbv4 [1] | 178/23/22 | 178/23/22 | 915/100/117 |
| NSCLCR [2] | 331/40/44 | 331/40/44 | 331/40/44 |
| LIDC-IDRI [3, 4] | 646/161/63 | 646/161/63 | 1899/469/216 |
| DLCS24 [5] | 1054/354/197 | 1054/354/197 | 1613/572/293 |
| IMD-CT [6] | 1549/192/195 | 1549/192/195 | 1549/192/195 |
| LUNA25 [7] | 1594/199/200 | 3039/389/364 | 4745/571/583 |
| **Total** | **5352/969/721** | **6797/1159/885** | **11052/1944/1448** |

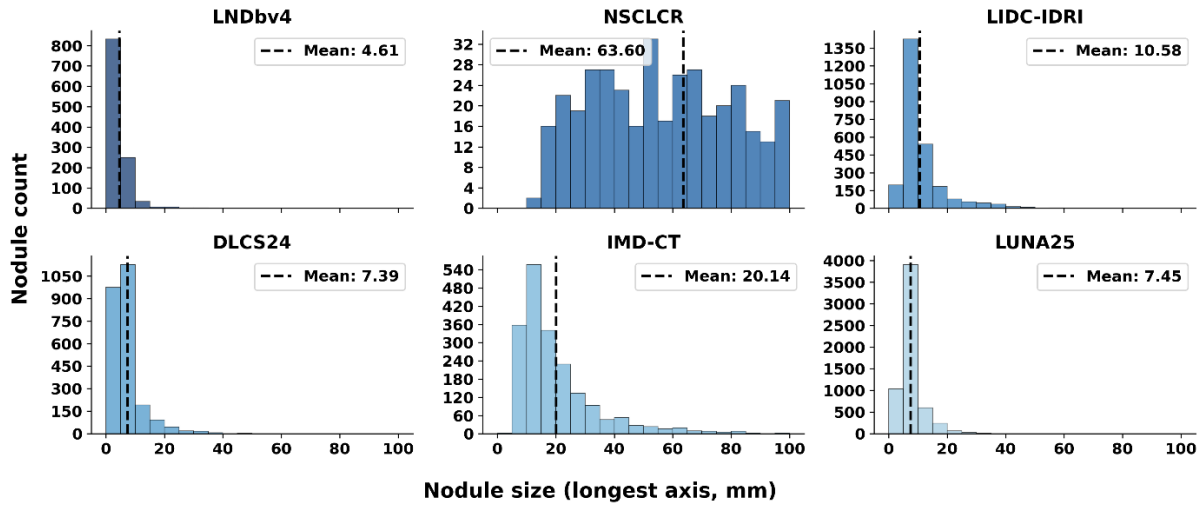

**Figure S1. Distribution of nodule sizes across datasets.** Histograms of nodule longest-axis diameter (mm) for each dataset (LNDbv4, NSCLCR, LIDC-IDRI, DLCS24, IMD-CT, and LUNA25). The dashed vertical line denotes the mean nodule size within each dataset, illustrating the substantial variation in size distributions across cohorts.

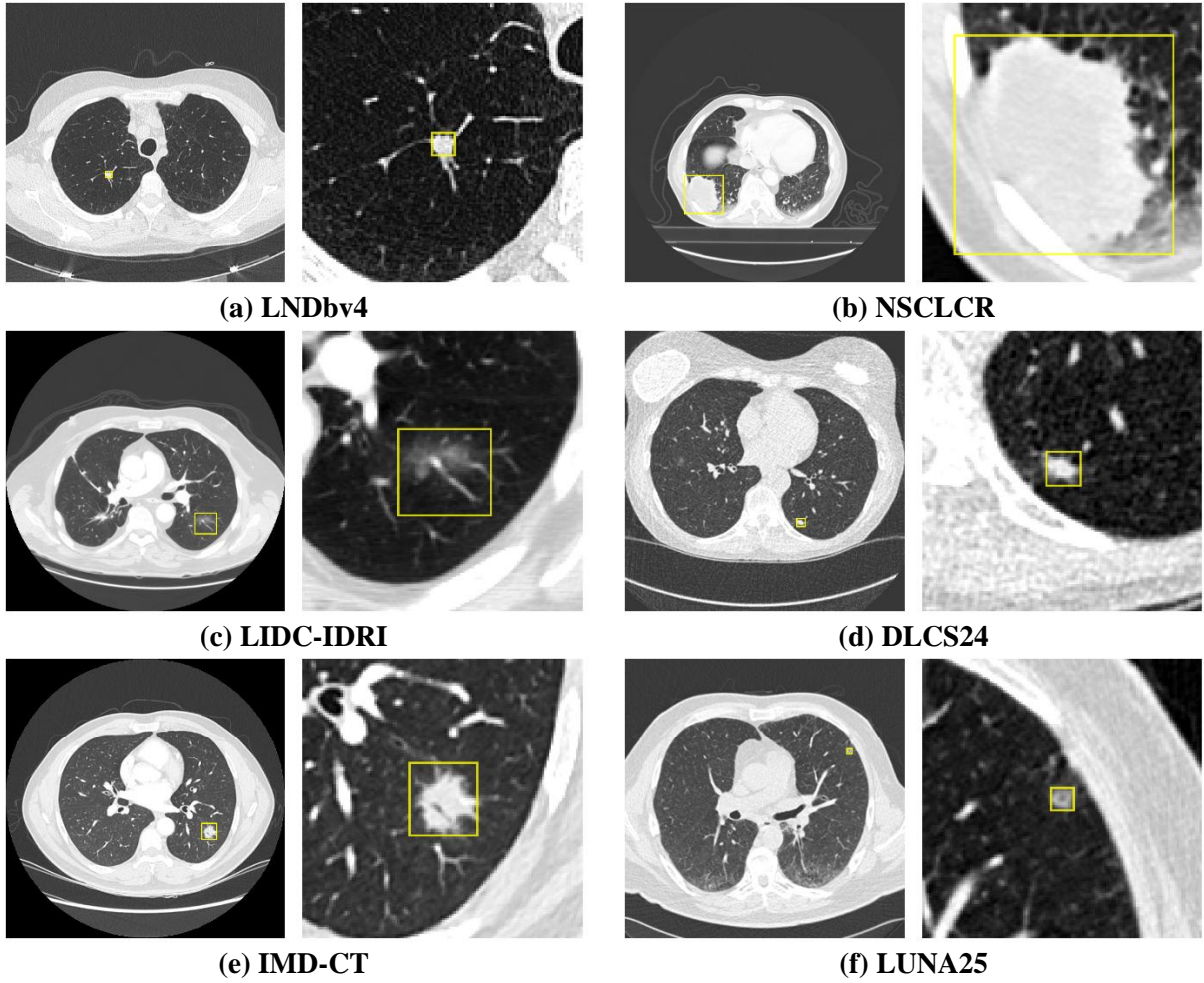

**Figure S2. Representative nodule examples across the six source datasets.** For each dataset, an axial CT slice is shown with a bounding box highlighting the target lesion, along with a corresponding zoomed-in view to illustrate variability in image quality, nodule size, texture, and surrounding anatomy: (a) LNDbv4, (b) NSCLC-R, (c) LIDC-IDRI, (d) DLCS24, (e) IMD-CT, and (f) LUNA25.

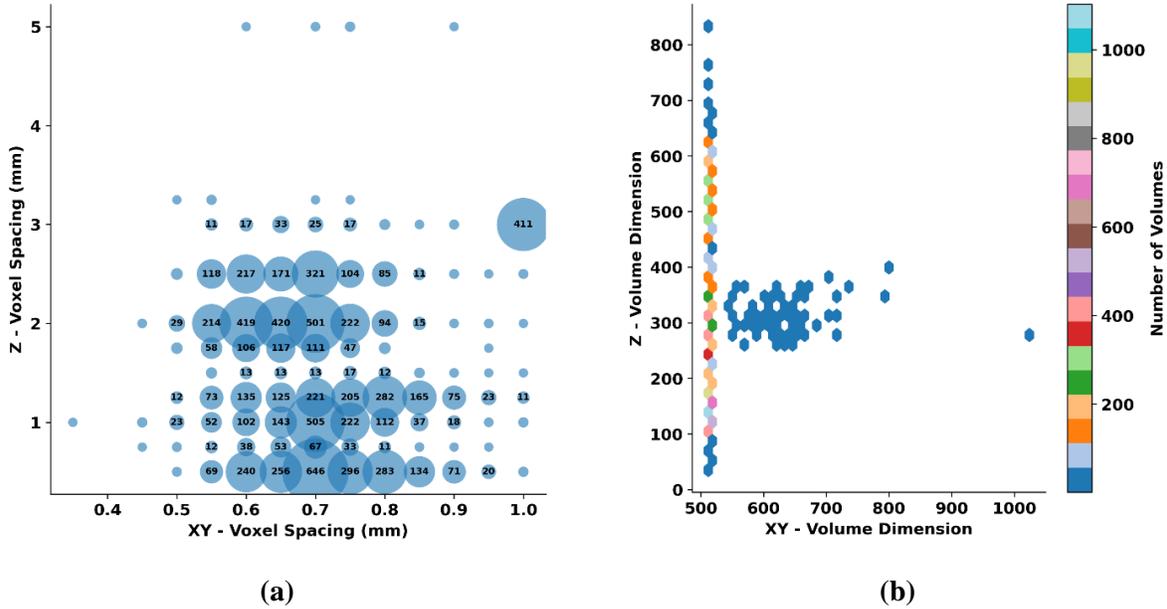

**Figure S2. Heterogeneity of CT resolutions and volume sizes in the combined cohort.**
**(a)** Bubble plot of in-plane (XY) versus through-plane (Z) voxel spacing (mm). Each circle represents one unique spacing configuration, and the circle area is proportional to the number of volumes with that spacing (numbers inside the circles indicate counts). **(b)** Scatter plot of in-plane matrix size (XY volume dimension) versus number of slices (Z volume dimension). The accompanying color bar indicates the number of volumes sharing a given combination of in-plane and through-plane dimensions.

### S1.1. Lung Nodule Database (LNDbv4) [1] :

**Table S2.** Patient Demographics and clinical characteristics of **LNDbv4** dataset. Categorical variables are reported as count (%), and percentages are relative to the total cohort

| Category | Value | Total | Train | Validation | Test |
|---|---|---|---|---|---|
| **Patient** | Patients | 223 (100.0%) | 178 (79.8%) | 23 (10.3%) | 22 (9.9%) |
| **CT** | CT scans | 1132 (100.0%) | 915 (80.8%) | 100 (8.8%) | 117 (10.3%) |
| **Nodule** | Annotation Count | 1132 (100.0%) | 915 (80.8%) | 100 (8.8%) | 117 (10.3%) |
| | Mean ± SD (mm) | 8.0 ± 5.2 | 8.2 ± 5.2 | 8.4 ± 6.2 | 6.2 ± 2.8 |
| **Diagnosis; Radiologist Suspicion Label (RSL)** | No-Cancer (0–<3) | 502 (87.3%) | 415 (87.4%) | 38 (79.2%) | 49 (94.2%) |
| | Cancer (4–5) | 73 (12.7%) | 60 (12.6%) | 10 (20.8%) | 3 (5.8%) |

## S1.2. NSCLC-Radiomics (NSCLC-R) Dataset [2]:

We utilized the non-small cell lung cancer (NSCLC) Radiomics (**NSCLC-Radiomics**) dataset [2, 8], comprising pre-treatment CT scans from 415 patients diagnosed with NSCLC. Each scan includes a **gross tumor volume (GTV-1)** manually delineated by a radiation oncologist. From these segmentation masks, we derived 3D bounding box annotations to localize tumor regions for analysis. Clinical metadata are available, including age, gender, TNM staging, overall stage, and histological subtype. For patients with missing histology, we assigned the label "**unknown**". During preprocessing, one case was excluded due to errors in the segmentation mask, resulting in a final cohort of 415 patients. The dataset was then stratified by histological subtype and split into training (80%), validation (10%), and test (10%) subsets. **Table S3** reports summary statistics for 415 NSCLC patients, including age (mean ± standard deviation), gender distribution, histological subtype, and staging.

**Table S3.** Patient Demographics and clinical characteristics of NSCLC-Radiomics dataset. Categorical variables are reported as count (%), and percentages are relative to the total cohort (N = 415).

| Category | Value | Total | Train | Validation | Test |
|---|---|---|---|---|---|
| **Patient** | Patients | 415 (100.0%) | 331 (79.76%) | 40 (9.6%) | 44 (10.6%) |
| **CT** | CT scans | 415 (100.0%) | 331 (79.76%) | 40 (9.6%) | 44 (10.6%) |
| **Age (years)** | Mean ± SD | 68.2 ± 10.1 | 68.1 ± 9.9 | 69.1 ± 10.6 | 67.9 ± 10.9 |
| **Gender** | Female | 128 (30.8%) | 110 (33.23%) | 10 (25.0%) | 8 (18.2%) |
| | Male | 287 (69.2%) | 221 (66.77%) | 30 (75.0%) | 36 (81.8%) |
| **Nodule** | Annotation Count | 415 (100.0%) | 331 (79.8%) | 40 (9.6%) | 44 (10.6%) |
| | Mean ± SD (mm) | 95.2 ± 45.6 | 95.9 ± 45.4 | 91.8 ± 45.2 | 93.1 ± 48.3 |
| **Overall Stage** | I | 92 (22.2%) | 76 (22.96%) | 6 (15.0%) | 10 (22.7%) |
| | II | 40 (9.6%) | 33 (9.97%) | 4 (10.0%) | 3 (6.8%) |
| | IIIa | 111 (26.8%) | 90 (27.2%) | 12 (30.0%) | 9 (20.5%) |
| | IIIb | 171 (41.2%) | 131 (39.58%) | 18 (45.0%) | 22 (50.0%) |
| **Histology** | Adenocarcinoma | 50 (12.1%) | 39 (11.78%) | 5 (12.50%) | 6 (13.6%) |
| | Large cell | 114 (27.5%) | 91 (27.49%) | 11 (27.5%) | 12 (27.8%) |
| | NOS | 61 (14.7%) | 49 (14.80%) | 5 (12.5%) | 7 (15.9%) |
| | Squamous cell carcinoma | 151 (36.4%) | 121 (36.6%) | 15 (37.5%) | 15 (34.1%) |
| | **Unknown** | 39 (9.4%) | 31 (9.4%) | 4 (10.0%) | 4 (9.1%) |

## S1.3. Lung Image Database Consortium and Image Database Resource Initiative (LIDC-IDRI) [3, 4]:

We used a curated subset of the Lung Image Database Consortium and Image Database Resource Initiative (LIDC–IDRI) [3, 4] comprising 870 patients (one CT per patient) interpreted by up to four thoracic radiologists. Reader-drawn nodule contours were merged with a voxel-wise union to obtain a single segmentation per lesion, yielding 2,584 annotated nodules with a mean size of 11.8 ± 10.4 mm. The dataset was split at the patient level into patients for train /validation/test **as summarized in Table S4**. For malignancy classification, we further used a previously defined subset of 677 nodules with **Radiologist Suspicion Labels (RSL)** (350 no-cancer, 327 cancer) [9, 10]. The RSL is a proxy malignancy standard derived from the radiologists' subjective assessments on LIDC–IDRI and, although it lacks histopathologic confirmation, has been widely adopted for benchmarking deep learning models on this dataset, similar to prior studies [9, 10]

**Table S4.** Patient Demographics and clinical characteristics of LIDC-IDRI dataset. Categorical variables are reported as count (%), and percentages are relative to the total cohort.

| Category | Value | Total | Train | Validation | Test |
|---|---|---|---|---|---|
| **Patient** | Patients | 870 (100.0%) | 646 (74.3%) | 161 (18.5%) | 63 (7.2%) |
| **CT** | CT scans | 870 (100.0%) | 646 (74.3%) | 161 (18.5%) | 63 (7.2%) |
| **Nodule** | Annotation Count | 2584 (100.0%) | 1899 (73.5%) | 469 (18.2%) | 216 (8.4%) |
| | Mean ± SD (mm) | 11.8 ± 10.4 | 11.9 ± 10.4 | 12.0 ± 10.5 | 10.4 ± 9.6 |
| **Radiologist Suspicion Label (RSL)** | | | | | |
| | No-Cancer | 350 (51.7) | | | 350 (51.7) |
| | Cancer | 327 (48.3) | | | 327 (48.3) |

**S1.4. Integrated Multiomics (IMD-CT) Dataset [6] :**

We used a multi-institutional dataset comprising 1936 patients with indeterminate pulmonary nodules (IPNs), as recently described in Zhao et al. [6]. Each patient had one annotated nodule and corresponding chest CT, with an average age of 56 ± 12.3 years. We randomly split the dataset into 80% for training, and 10% each for validation and testing, ensuring balanced distributions of cancer labels and patient characteristics across the subsets. A total of 19.9% of nodules were histopathological confirmed benign, and 80.1% malignant. A detailed summary of patient and nodule characteristics across the training, validation, and test subsets is provided in **Table S5**.

**Table S5.** Summary of patient, CT scan, and lesion characteristics across dataset splits (training, validation, test). Values are shown as count (percentage) for categorical variables and mean ± standard deviation for continuous variables.

| Category | Value | Total | Train | Validation | Test |
|---|---|---|---|---|---|
| **Patient** | Patients | 1936 (100.0%) | 1549 (80.0%) | 192 (9.9%) | 195 (10.1%) |
| **CT** | CT scans | 1936 (100.0%) | 1549 (80.0%) | 192 (9.9%) | 195 (10.1%) |
| **Age (years)** | Mean ± SD | 56.0 ± 12.3 | 55.8 ± 12.5 | 57.6 ± 11.0 | 55.9 ± 11.5 |
| **Gender** | Female | 1164 (60.1%) | 953 (61.5%) | 95 (49.5%) | 116 (59.5%) |
| | Male | 772 (39.9%) | 596 (38.5%) | 97 (50.5%) | 79 (40.5%) |
| **Emphysema** | No | 1694 (87.5%) | 1362 (87.9%) | 167 (87.0%) | 165 (84.6%) |
| | Yes | 242 (12.5%) | 187 (12.1%) | 25 (13.0%) | 30 (15.4%) |
| **Smoking History** | Ever | 534 (27.6%) | 424 (27.4%) | 53 (27.6%) | 57 (29.2%) |
| | Never | 1402 (72.4%) | 1125 (72.6%) | 139 (72.4%) | 138 (70.8%) |
| **Nodule** | Annotation Count | 1936 (100.0%) | 1549 (80.0%) | 192 (9.9%) | 195 (10.1%) |
| | Mean ± SD (mm) | 30.2 ± 20.8 | 29.8 ± 20.5 | 33.9 ± 23.6 | 29.3 ± 20.3 |
| **Diagnosis** | Benign | 386 (19.9%) | 311 (20.1%) | 37 (19.3%) | 38 (19.5%) |
| | Cancer | 1550 (80.1%) | 1238 (79.9%) | 155 (80.7%) | 157 (80.5%) |

## S1.5. Duke Lung Cancer Screening Dataset 2024 [5]:

**Table S6. DLCS24 cohort characteristics and data splits.** Summary of patient- and nodule-level statistics for the DLCS24 dataset and the train/validation/test partitions, including number of patients and CT scans, age (mean ± SD), sex distribution, total nodule annotation count and mean nodule size (longest-axis diameter, mm), and diagnosis labels (benign vs cancer). Values are reported as n (%) unless otherwise specified.

| Category | Value | Total | Train | Validation | Test |
|---|---|---|---|---|---|
| Patient | Patients | 1605 (100.0%) | 1054 (65.7%) | 354 (22.1%) | 197 (12.3%) |
| CT | CT scans | 1605 (100.0%) | 1054 (65.7%) | 354 (22.1%) | 197 (12.3%) |
| Age (years) | Mean ± SD | 66.8 ± 6.1 | 66.9 ± 6.2 | 66.4 ± 6.0 | 67.0 ± 5.9 |
| Gender | Female | 1251 (50.5%) | 776 (48.1%) | 304 (53.1%) | 171 (58.4%) |
|  | Male | 1227 (49.5%) | 837 (51.9%) | 268 (46.9%) | 122 (41.6%) |
| Nodule | Annotation Count | 2478 (100.0%) | 1613 (65.1%) | 572 (23.1%) | 293 (11.8%) |
|  | Mean ± SD (mm) | 12.3 ± 8.9 | 12.6 ± 9.6 | 11.8 ± 7.6 | 11.7 ± 7.0 |
| Diagnosis | Benign | 2214 (89.3%) | 1447 (89.7%) | 507 (88.6%) | 260 (88.7%) |
|  | Cancer | 264 (10.7%) | 166 (10.3%) | 65 (11.4%) | 33 (11.3%) |

## S1.6. LUNA25 [7]:

**Table S7. LUNA25 cohort characteristics and train/validation/test splits.** Counts of patients, CT scans, and nodule annotations with age (mean ± SD), sex distribution, mean nodule size (mm), and diagnosis labels summarized for the full cohort and each split.

| Category | Value | Total | Train | Validation | Test |
|---|---|---|---|---|---|
| Patient | Patients | 1993 (100.0%) | 1594 (80.0%) | 199 (10.0%) | 200 (10.0%) |
| CT | CT scans | 3792 (100.0%) | 3039 (80.1%) | 389 (10.3%) | 364 (9.6%) |
| Age (years) | Mean ± SD | 63.3 ± 5.3 | 63.4 ± 5.3 | 63.0 ± 4.9 | 63.4 ± 5.6 |
| Gender | Female | 2414 (40.9%) | 1941 (40.9%) | 216 (37.8%) | 257 (44.1%) |
|  | Male | 3485 (59.1%) | 2804 (59.1%) | 355 (62.2%) | 326 (55.9%) |
| Nodule | Annotation Count | 5899 (100.0%) | 4745 (80.4%) | 571 (9.7%) | 583 (9.9%) |
|  | Mean ± SD (mm) | 12.7 ± 6.8 | 12.7 ± 6.9 | 13.1 ± 6.4 | 12.5 ± 6.6 |
| Diagnosis | Benign | 5450 (92.4%) | 4393 (92.6%) | 523 (91.6%) | 534 (91.6%) |
|  | Cancer | 449 (7.6%) | 352 (7.4%) | 48 (8.4%) | 49 (8.4%) |

# S2. Annotation workflow

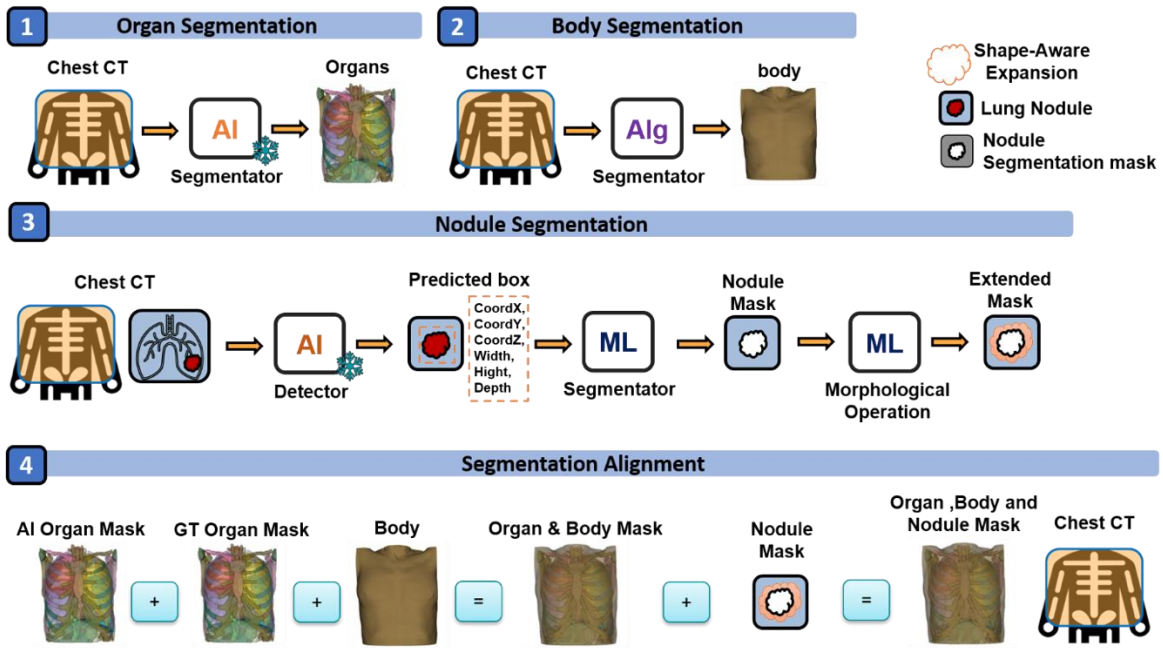

**Figure S4. Workflow for constructing the NodMAISI development dataset.** The pipeline includes (1) organ segmentation using AI models, (2) body segmentation with algorithmic methods, (3) nodule segmentation through AI-assisted and ML-based refinement, and (4) segmentation alignment to integrate organs, body, and nodules segmentations into anatomically consistent volumes.

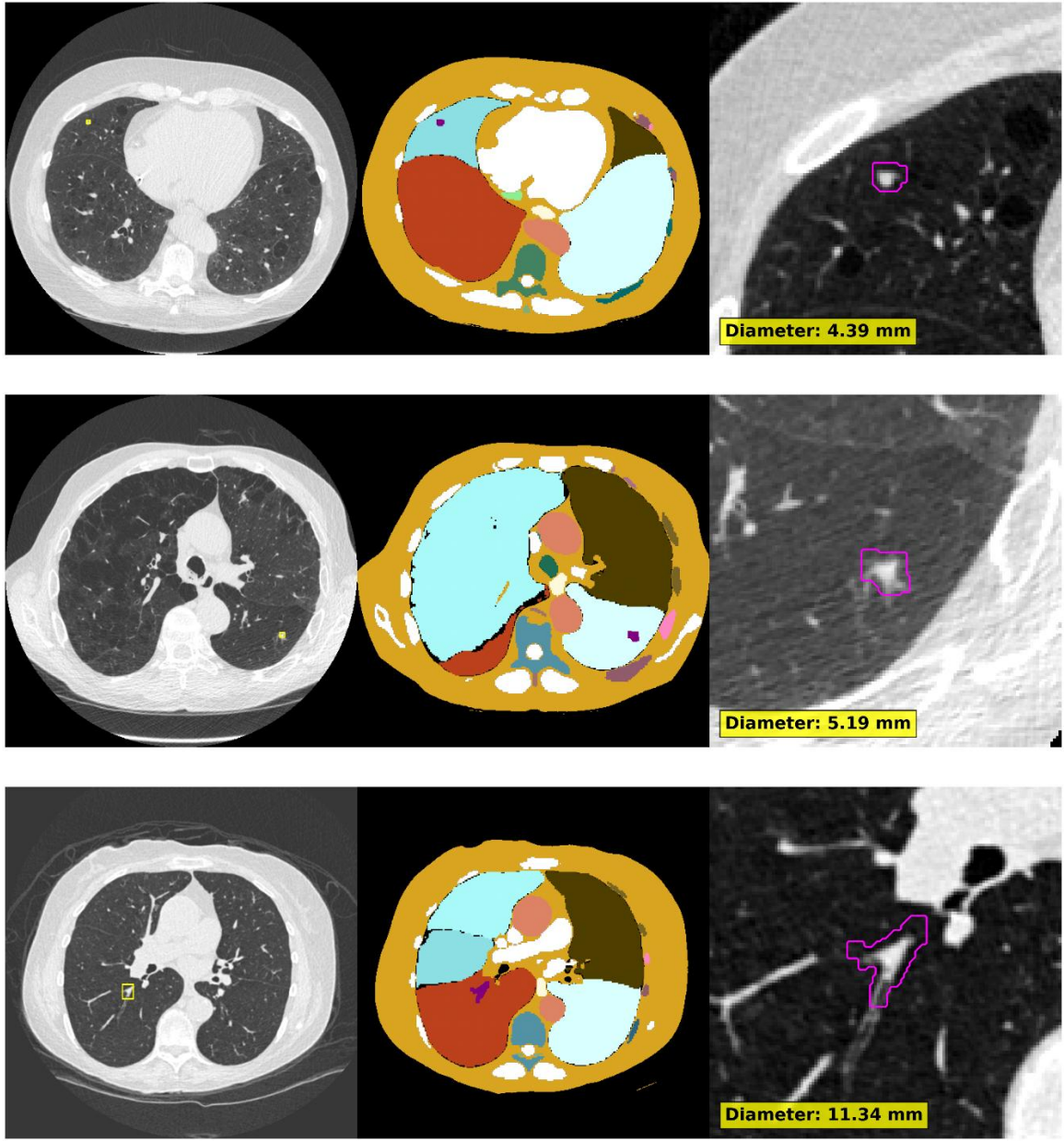

**Figure S5.** Qualitative examples of the standardized segmentation outputs. For representative clinical CT slices (left), we show the corresponding automatically generated multi-organ/body segmentation masks (middle) and a zoomed-in view of the refined nodule segmentation contour (right, magenta). The yellow annotation indicates the estimated nodule diameter (mm) for each example.

## S3. Pseudo Nodule Segmentation:

**Segmentation Algorithm:** We designed a point-driven, weakly supervised 3D nodule segmentation framework that leverages bounding-box priors to localize the candidate region. For each annotated lesion, a cubic volume-of-interest (VOI) was extracted around the bounding-box center, extending the box dimensions by a fixed margin to incorporate perinodular context. Within this VOI, segmentation was performed using unsupervised intensity-based methods, including k-means clustering, Gaussian Mixture Models (GMM), fuzzy c-means (FCM), and Otsu thresholding. The predicted mask was then refined with 3D morphological closing and opening to suppress noise and small gaps, and finally mapped back to the global CT space. Optional modules allowed expansion of the mask by a user-defined physical distance (mm) and determination of lobar location and pleural distance using lung-lobe masks and distance transforms. This modular design enabled consistent segmentation from bounding-box inputs without requiring voxel-level annotations.

**K-means Variant:** In the k-means implementation, the algorithm first extracts a local 3D patch centered on the bounding-box coordinates of the candidate nodule, extending the box boundaries by a fixed margin (default: 5 voxels) to provide contextual tissue. This patch is reshaped into a one-dimensional intensity vector and clustered into n_clusters = 2 groups, with the assumption that one cluster corresponds to lung parenchyma/background and the other to the higher-attenuation nodule. The nodule cluster is automatically identified as the one with the higher mean Hounsfield unit value, which is most consistent with solid or part-solid nodules. The resulting binary mask is reshaped back into the 3D VOI and refined using binary closing and binary opening (structuring elements of $3^3$ and $2^3$ voxels, respectively) to smooth boundaries and suppress isolated noise. The cleaned mask is finally placed back into the global CT volume at the appropriate coordinates.

**Key parameters:**

- bbox_center (x, y, z) and bbox_whd (width, height, depth): define the candidate's bounding box.
- margin (default 5): extra voxels added around the box to incorporate perinodular tissue.
- n_clusters (default 2): number of k-means clusters; increasing this could capture more tissue types but risks oversegmentation.
- random_state (0): ensures reproducibility of clustering.

**Evaluation Criteria:** In the absence of voxel-level ground truth, we adopted a bounding box–supervised evaluation strategy to assess segmentation performance. Each CT volume was accompanied by annotations specifying the nodule center in world coordinates and its dimensions in millimeters, which were converted into voxel indices using the image spacing and clipped to the volume boundaries. A binary mask representing the bounding box was then constructed and used as a weak surrogate for ground truth. we extracted a patch centered on the bounding box, extending it by a fixed margin (**64 voxels**) to define the volume of interest (VOI). Predicted segmentation masks were cropped to the same VOI-constrained region of interest, and performance was quantified in terms of **Dice similarity coefficient**. Metrics were computed per lesion. This evaluation strategy enables consistent comparison of segmentation algorithms under weak supervision while acknowledging the limitations of not having voxel-level annotations.

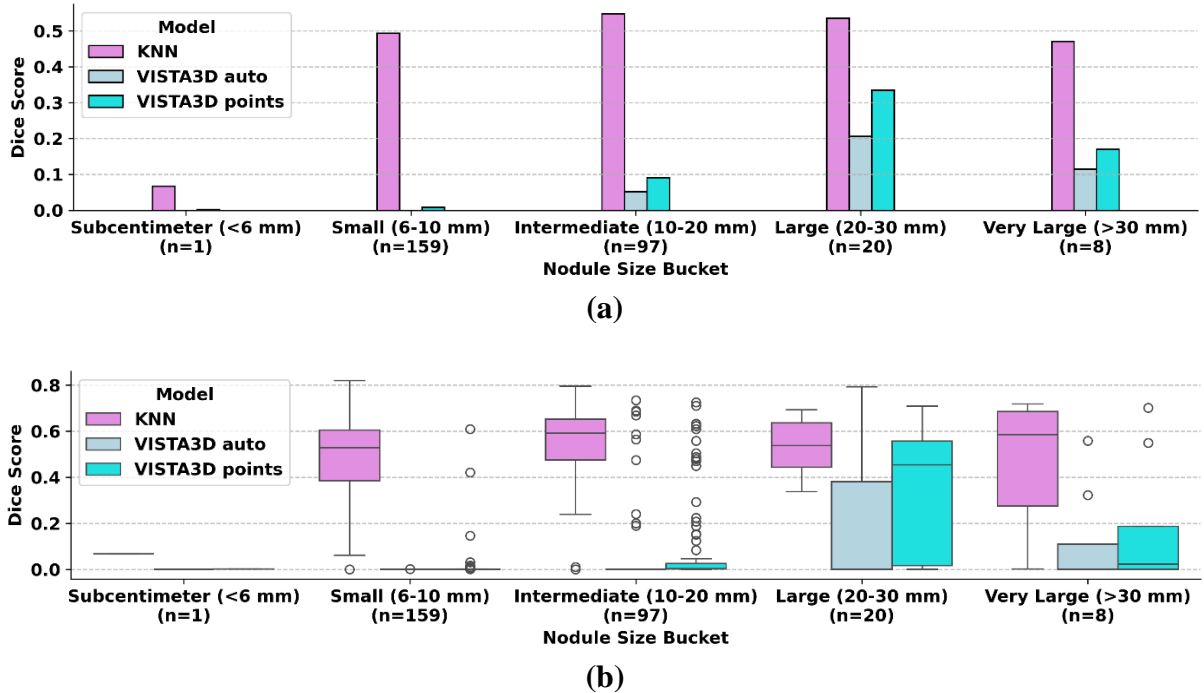

**Figure S6.** Segmentation performance of KNN, VISTA3D auto, and VISTA3D points across different nodule size buckets. (A) Bar plots display the mean Dice similarity coefficient for each model and size category. (B) Boxplots show the distribution of Dice scores, with boxes representing the interquartile range, horizontal lines indicating the median, whiskers extending to 1.5× the interquartile range, and circles denoting outliers.

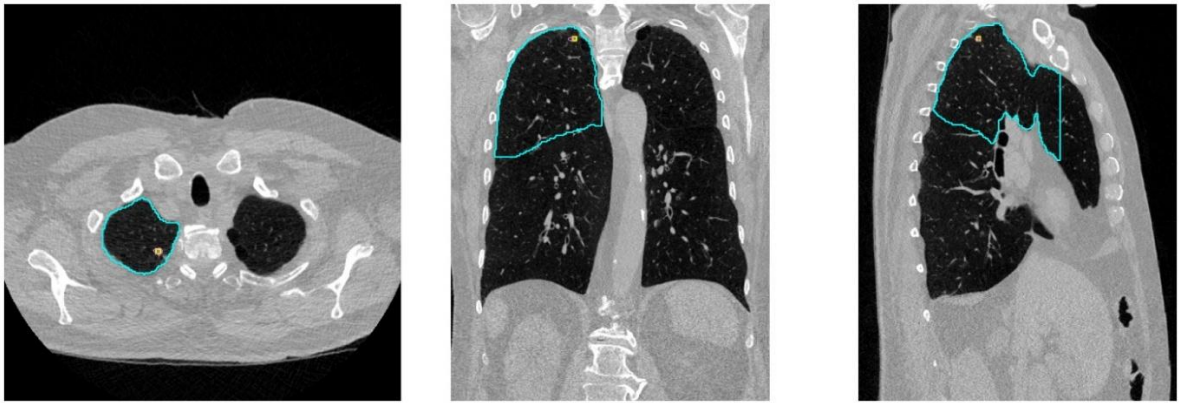

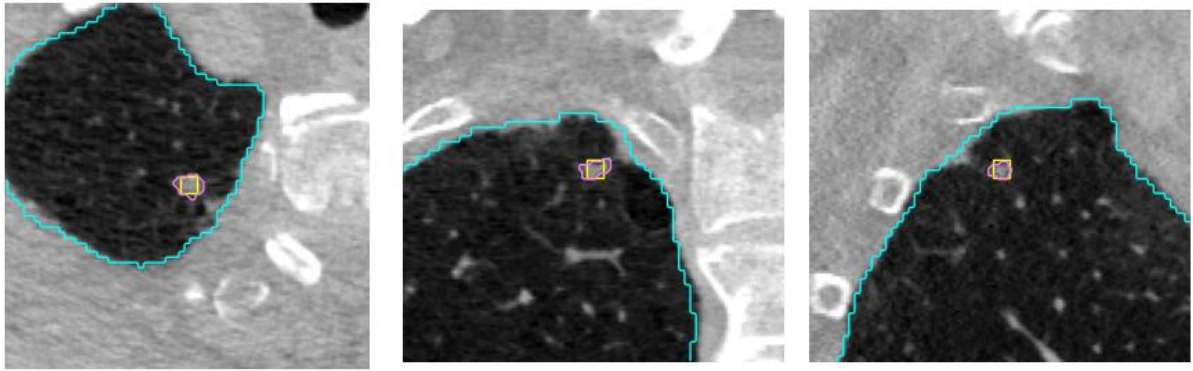

**Figure S7.** Segmentation visualization of a right upper-lobe nodule (**largest axis 4.5 mm**) in a 77-year-old male (Lung-RADS 3.0, benign diagnosis). Axial, coronal, and sagittal CT views are displayed. The violet contour represents the K-means (KNN)–based segmentation, the cyan contour denotes the VISTA3D point-driven method, and the light blue contour corresponds to the full VISTA3D segmentation, which failed to capture the nodule in this case.

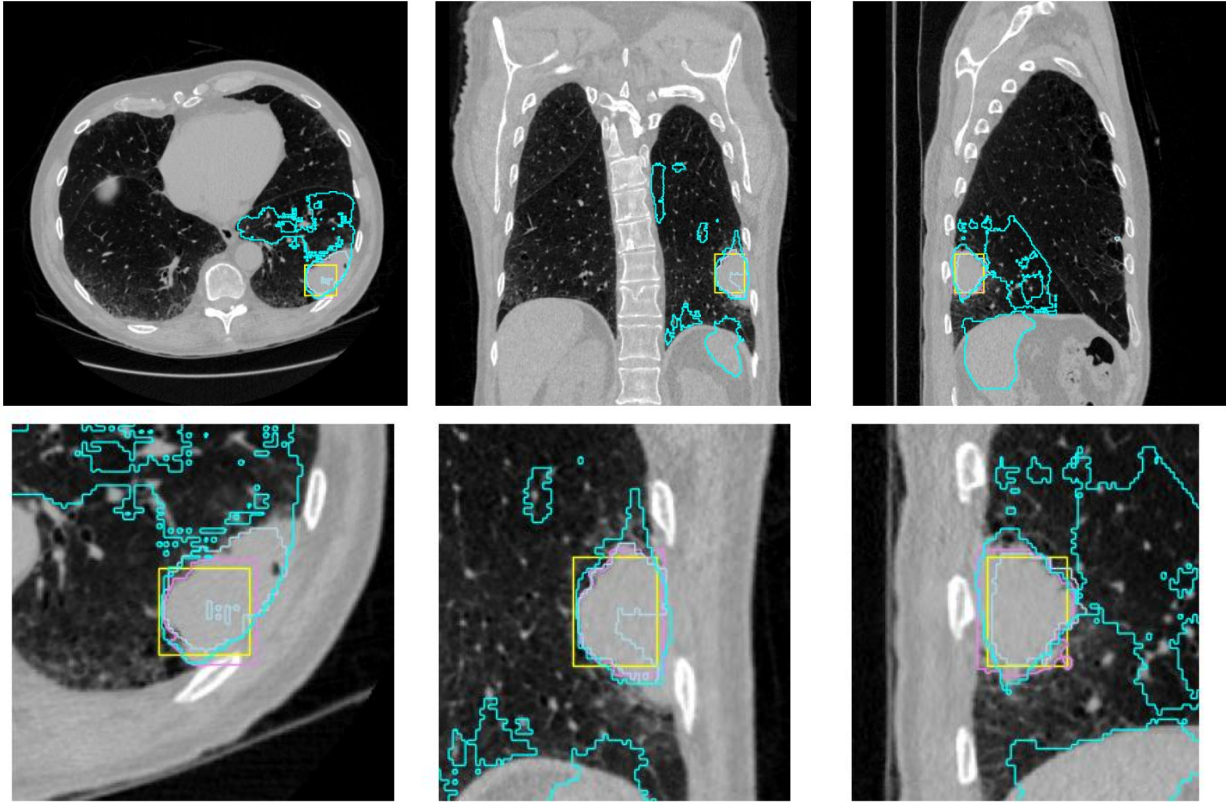

**Figure S8.** Example of a malignant pulmonary nodule annotation from the DLCS cohort. Axial (left), coronal (middle), and sagittal (right) CT views (top row) show the annotated lesion in the **left lower lobe**, with bounding boxes and contour overlays. The bottom row provides zoomed-in views of the nodule. The **cyan contour** denotes the reference segmentation, the **purple outline** marks the lesion core, and the **yellow box** shows the candidate bounding region. This case corresponds to a **70-year-old Caucasian male (not Hispanic/Latino)** with a **31.8 × 31.2 × 33.2 mm lung nodule**, located **1.3 mm from the pleural surface**, scored as **Lung-RADS 5**, and confirmed as **cancer-positive**.

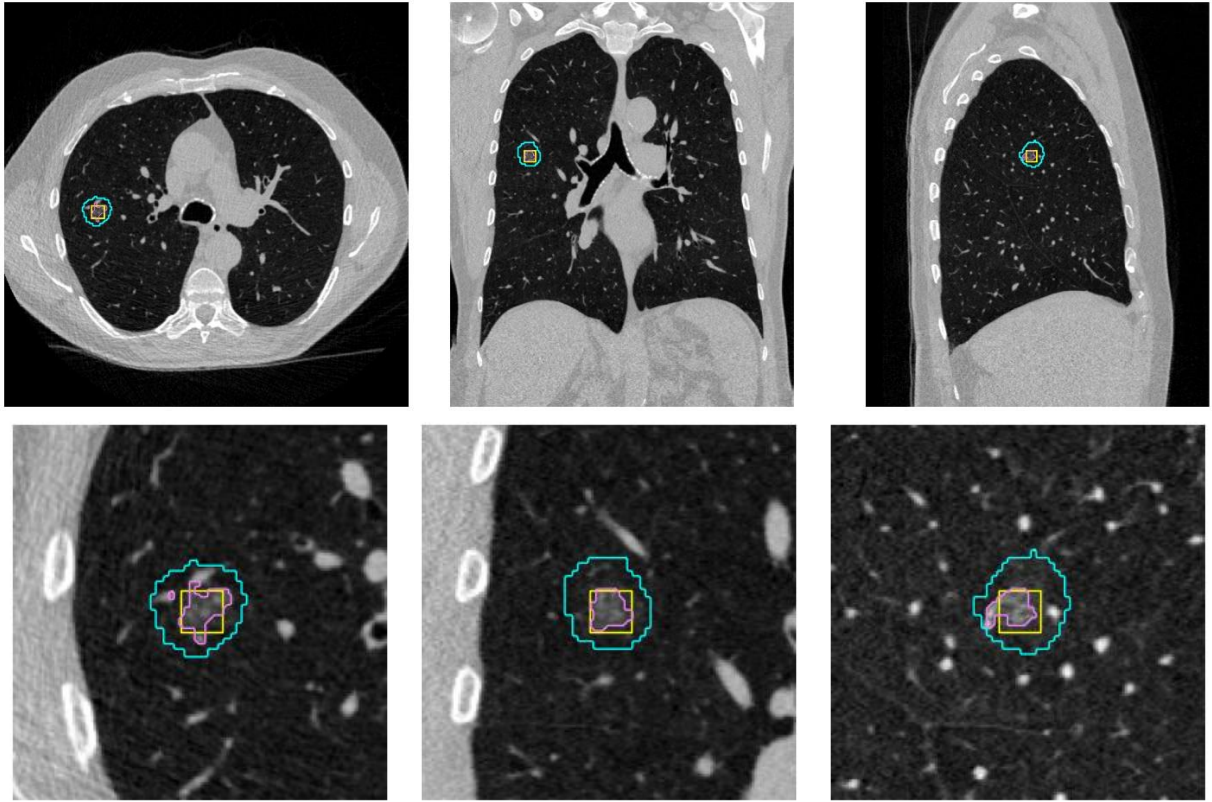

**Figure S9.** Example of a malignant pulmonary nodule annotation from the DLCS cohort. Shown are axial (left), coronal (middle), and sagittal (right) CT views (top row) with bounding box overlays highlighting the candidate lesion, and corresponding zoomed-in views (bottom row) with detailed contours. The cyan outline denotes the reference segmentation boundary, the purple contour indicates lesion core annotation, and the yellow bounding box corresponds to the candidate detection region. This case is a **74-year-old Caucasian male (not Hispanic/Latino)**, with a **right upper lobe nodule** measuring approximately **11.4 × 11.4 × 10.8 mm**, located **53.5 mm from the lung wall**, scored as **Lung-RADS 6**, and confirmed as **cancer-positive**.

# S4. Results: Lesion Detectability

**Table S8. Size-stratified lesion detectability (FROC) on IMD-CT test set.**
Sensitivity is reported at average false positives per scan (FP/scan) thresholds (1/8, 1/4, 1/2, 1, 2, 4, 8) for **Clinical**, **MAISI-v2**, and **NodMAISI**. "Avg" denotes mean sensitivity across the FP/scan operating points. "Detection rate" reports the fraction of detected nodules (Detected/Total) in each size bin (<10 mm, 10–<20 mm, ≥20 mm) and overall.

| Category | Model | Sensitivity at Avg. FP/Scans | | | | | | | Detection rate (Detected/Total) |
| --- | --- | --- | --- | --- | --- | --- | --- | --- | --- |
| | | Avg | @1/8 | @1/4 | @1/2 | @1 | @2 | @4 | @8 | |
| **Size bin** | | | | | | | | | | |
| <10 | Clinical | **0.82** | 0.40 | 0.67 | 0.77 | 0.91 | 0.98 | 1.00 | 1.00 | 1.00 (43/43) |
| | MAISI-v2 | 0.05 | 0.05 | 0.05 | 0.05 | 0.05 | 0.05 | 0.05 | 0.05 | 0.05 (2/43) |
| | NodMAISI | 0.45 | 0.33 | 0.37 | 0.37 | 0.42 | 0.49 | 0.56 | 0.60 | 0.70 (30/43) |
| **10 to <20** | Clinical | 0.68 | 0.36 | 0.43 | 0.54 | 0.72 | 0.86 | 0.91 | 0.97 | 0.98 (88/90) |
| | MAISI-v2 | 0.40 | 0.31 | 0.33 | 0.37 | 0.41 | 0.44 | 0.46 | 0.49 | 0.50 (45/90) |
| | NodMAISI | **0.74** | 0.57 | 0.63 | 0.69 | 0.77 | 0.82 | 0.83 | 0.86 | 0.90 (81/90) |
| **>=20** | Clinical | 0.61 | 0.35 | 0.35 | 0.44 | 0.56 | 0.77 | 0.84 | 0.94 | 0.97 (60/62) |
| | MAISI-v2 | 0.59 | 0.34 | 0.42 | 0.48 | 0.61 | 0.68 | 0.76 | 0.87 | 0.89 (55/62) |
| | NodMAISI | **0.79** | 0.69 | 0.73 | 0.74 | 0.79 | 0.85 | 0.85 | 0.90 | 0.92 (57/62) |
| **Overall** | Clinical | 0.68 | 0.35 | 0.43 | 0.56 | 0.69 | 0.84 | 0.92 | 0.96 | 0.98 (191/195) |
| | MAISI-v2 | 0.39 | 0.28 | 0.31 | 0.34 | 0.41 | 0.44 | 0.46 | 0.51 | 0.52 (102/195) |
| | NodMAISI | **0.69** | 0.54 | 0.60 | 0.64 | 0.70 | 0.76 | 0.78 | 0.81 | 0.86 (168/195) |

**Table S9. Size-stratified lesion detectability (FROC) on DLCS24 test set.**
Sensitivity is reported at average FP/scan thresholds (1/8, 1/4, 1/2, 1, 2, 4, 8) for **Clinical**, **MAISI-v2**, and **NodMAISI**. "Avg" denotes mean sensitivity across the FP/scan operating points. "Detection rate" reports the fraction of detected nodules (Detected/Total) in each size bin and overall.

| Category | Model | Sensitivity at Avg. FP/Scans | | | | | | | Detection rate (Detected/Total) |
| --- | --- | --- | --- | --- | --- | --- | --- | --- | --- |
| | | Avg | @1/8 | @1/4 | @1/2 | @1 | @2 | @4 | @8 | |
| **Size bin** | | | | | | | | | | |
| **<10** | Clinical | 0.56 | 0.15 | 0.23 | 0.39 | 0.60 | 0.74 | 0.86 | 0.95 | 0.97 (248/256) |
| | MAISI-v2 | 0.15 | 0.09 | 0.11 | 0.12 | 0.15 | 0.18 | 0.21 | 0.23 | 0.25 (63/256) |
| | NodMAISI | 0.61 | 0.36 | 0.48 | 0.59 | 0.66 | 0.71 | 0.73 | 0.76 | 0.85 (217/256) |
| **10 to <20** | Clinical | 0.51 | 0.14 | 0.34 | 0.34 | 0.38 | 0.66 | 0.76 | 0.93 | 0.93 (27/29) |
| | MAISI-v2 | 0.51 | 0.24 | 0.34 | 0.38 | 0.52 | 0.59 | 0.69 | 0.83 | 0.93 (27/29) |
| | NodMAISI | 0.58 | 0.21 | 0.34 | 0.59 | 0.66 | 0.72 | 0.76 | 0.76 | 0.93 (27/29) |
| **>=20** | Clinical | 0.32 | 0.12 | 0.12 | 0.12 | 0.12 | 0.50 | 0.50 | 0.75 | 0.88 (7/8) |
| | MAISI-v2 | 0.27 | 0.00 | 0.00 | 0.12 | 0.25 | 0.38 | 0.50 | 0.62 | 0.75 (6/8) |
| | NodMAISI | 0.54 | 0.38 | 0.38 | 0.38 | 0.62 | 0.62 | 0.62 | 0.75 | 0.88 (7/8) |
| **Overall** | Clinical | 0.57 | 0.19 | 0.28 | 0.42 | 0.59 | 0.74 | 0.84 | 0.95 | 0.96 (282/293) |
| | MAISI-v2 | 0.20 | 0.11 | 0.14 | 0.15 | 0.19 | 0.22 | 0.27 | 0.30 | 0.33 (96/293) |
| | NodMAISI | 0.63 | 0.44 | 0.54 | 0.59 | 0.68 | 0.71 | 0.73 | 0.76 | 0.86 (251/293) |

# S5. Results: Downstream Classification Performance

**Table S10.** External test Area under the ROC curve (AUC) and 95% confidence intervals (95% CI) on **LUNA16** for clinical and generative-augmented models across clinical training data percentages. AUC and 95% CI on the external **LUNA16** are reported for each model instance trained on a given cross-validation fold and clinical training fraction (100%, 50%, 20%, 10%). Results are shown for the clinical-only model trained on LUNA25, the clinical model augmented with MAISI-v2 synthetic CT (Clinical + MAISI-v2), and the clinical model augmented with NodMAISI synthetic CT (Clinical + NodMAISI).
**Note:** "—" indicates model configurations that were not trained/evaluated at that training fraction.

|  | Clinical training dataset percentage | | | |
| --- | --- | --- | --- | --- |
| Fold | 100% | 50% | 20% | 10% |
| **Clinical (LUNA25)** | | | | |
| Fold 1 | 0.866 (0.839–0.893) | 0.799 (0.764–0.831) | 0.795 (0.759–0.828) | 0.746 (0.710–0.782) |
| Fold 2 | — | 0.828 (0.795–0.860) | 0.643 (0.600–0.686) | 0.574 (0.532–0.617) |
| Fold 3 | — | — | 0.739 (0.700–0.774) | 0.563 (0.519–0.607) |
| Fold 4 | — | — | 0.731 (0.693–0.770) | 0.574 (0.530–0.615) |
| Fold 5 | — | — | 0.697 (0.661–0.735) | 0.489 (0.447–0.533) |
| Fold 6 | — | — | — | 0.541 (0.497–0.585) |
| Fold 7 | — | — | — | 0.541 (0.498–0.584) |
| Fold 8 | — | — | — | 0.688 (0.647–0.727) |
| Fold 9 | — | — | — | 0.531 (0.489–0.573) |
| Fold 10 | — | — | — | 0.659 (0.615–0.701) |
| **Mean** | **0.867** | 0.814 | 0.721 | 0.590 |
| **Clinical + MAISI-v2** | | | | |
| Fold 1 | 0.858 (0.827–0.886) | 0.848 (0.816–0.878) | 0.813 (0.779–0.843) | 0.771 (0.735–0.807) |
| Fold 2 | — | 0.807 (0.773–0.838) | 0.786 (0.750–0.819) | 0.797 (0.761–0.830) |
| Fold 3 | — | — | 0.758 (0.719–0.794) | 0.755 (0.719–0.792) |
| Fold 4 | — | — | 0.786 (0.753–0.818) | 0.778 (0.744–0.813) |
| Fold 5 | — | — | 0.789 (0.752–0.823) | 0.736 (0.698–0.773) |
| Fold 6 | — | — | — | 0.753 (0.715–0.789) |
| Fold 7 | — | — | — | 0.781 (0.744–0.813) |
| Fold 8 | — | — | — | 0.777 (0.741–0.809) |
| Fold 9 | — | — | — | 0.745 (0.708–0.781) |
| Mean | 0.858 | 0.828 | 0.787 | 0.767 |
| **Clinical + NodMAISI** | | | | |
| Fold 1 | 0.861 (0.833–0.888) | 0.882 (0.854–0.908) | 0.845 (0.815–0.872) | 0.753 (0.718–0.788) |
| Fold 2 | — | 0.842 (0.814–0.872) | 0.834 (0.802–0.863) | 0.802 (0.769–0.834) |
| Fold 3 | — | — | 0.739 (0.701–0.773) | 0.754 (0.720–0.789) |
| Fold 4 | — | — | 0.785 (0.749–0.819) | 0.821 (0.788–0.849) |
| Fold 5 | — | — | 0.827 (0.795–0.856) | 0.815 (0.782–0.844) |
| Fold 6 | — | — | — | 0.746 (0.710–0.782) |
| Fold 7 | — | — | — | 0.826 (0.793–0.854) |
| Fold 8 | — | — | — | 0.822 (0.788–0.851) |
| Fold 9 | — | — | — | 0.818 (0.787–0.848) |
| Fold 10 | — | — | — | 0.804 (0.772–0.835) |
| **Mean** | **0.861** | **0.863** | **0.806** | **0.796** |

**Table S11.** External test Area under the ROC curve (AUC) and 95% confidence intervals (95% CI) on **LNDbv4** for clinical and generative-augmented models across clinical training data percentages. AUC and 95% CI on the external **LNDbv4** are reported for each model instance trained on a given cross-validation fold and clinical training fraction (100%, 50%, 20%, 10%). Results are shown for the clinical-only model trained on LUNA25, the clinical model augmented with MAISI-v2 synthetic CT (Clinical + MAISI-v2), and the clinical model augmented with NodMAISI synthetic CT (Clinical + NodMAISI). Note: "—" indicates model configurations that were not trained/evaluated at that training fraction.

| Fold | Clinical training dataset percentage | | | |
|---|---|---|---|---|
| | 100% | 50% | 20% | 10% |
| | **Clinical (LUNA25)** | | | |
| Fold 1 | 0.734 (0.663–0.801) | 0.742 (0.673–0.806) | 0.765 (0.705–0.822) | 0.736 (0.669–0.801) |
| Fold 2 | — | 0.720 (0.650–0.784) | 0.557 (0.493–0.619) | 0.628 (0.563–0.694) |
| Fold 3 | — | — | 0.681 (0.617–0.737) | 0.614 (0.547–0.676) |
| Fold 4 | — | — | 0.714 (0.657–0.765) | 0.516 (0.442–0.585) |
| Fold 5 | — | — | 0.617 (0.547–0.684) | 0.534 (0.470–0.598) |
| Fold 6 | — | — | — | 0.592 (0.517–0.658) |
| Fold 7 | — | — | — | 0.671 (0.613–0.731) |
| Fold 8 | — | — | — | 0.682 (0.620–0.741) |
| Fold 9 | — | — | — | 0.598 (0.528–0.662) |
| Fold 10 | — | — | — | 0.656 (0.595–0.717) |
| Mean | 0.734 | 0.731 | 0.666 | 0.622 |
| | **Clinical + MAISI-v2** | | | |
| Fold 1 | 0.803 (0.745–0.854) | 0.772 (0.710–0.829) | 0.744 (0.679–0.804) | 0.692 (0.622–0.757) |
| Fold 2 | — | 0.744 (0.675–0.807) | 0.762 (0.707–0.814) | 0.779 (0.720–0.831) |
| Fold 3 | — | — | 0.751 (0.698–0.801) | 0.717 (0.654–0.783) |
| Fold 4 | — | — | 0.725 (0.665–0.781) | 0.752 (0.686–0.811) |
| Fold 5 | — | — | 0.775 (0.721–0.825) | 0.723 (0.671–0.776) |
| Fold 6 | — | — | — | 0.740 (0.681–0.796) |
| Fold 7 | — | — | — | 0.749 (0.679–0.811) |
| Fold 8 | — | — | — | 0.741 (0.681–0.799) |
| Fold 9 | — | — | — | 0.745 (0.689–0.799) |
| Fold 10 | — | — | — | 0.732 (0.669–0.790) |
| Mean | **0.802** | 0.757 | 0.750 | 0.736 |
| | **Clinical + NodMAISI** | | | |
| Fold 1 | 0.794 (0.728–0.851) | 0.818 (0.764–0.867) | 0.830 (0.782–0.876) | 0.743 (0.676–0.803) |
| Fold 2 | — | 0.800 (0.745–0.854) | 0.765 (0.705–0.821) | 0.775 (0.718–0.827) |
| Fold 3 | — | — | 0.786 (0.729–0.843) | 0.762 (0.706–0.815) |
| Fold 4 | — | — | 0.762 (0.707–0.815) | 0.770 (0.714–0.821) |
| Fold 5 | — | — | 0.782 (0.724–0.837) | 0.791 (0.735–0.841) |
| Fold 6 | — | — | — | 0.746 (0.686–0.802) |
| Fold 7 | — | — | — | 0.788 (0.733–0.841) |
| Fold 8 | — | — | — | 0.798 (0.742–0.849) |
| Fold 9 | — | — | — | 0.804 (0.753–0.851) |
| Fold 10 | — | — | — | 0.787 (0.733–0.839) |
| Mean | 0.793 | **0.808** | **0.784** | **0.776** |

**Table S12.** External test Area under the ROC curve (AUC) and 95% confidence intervals (95% CI) on **DLCS24** for clinical and generative-augmented models across clinical training data percentages. AUC and 95% CI on the external **DLCS24** are reported for each model instance trained on a given cross-validation fold and clinical training fraction (100%, 50%, 20%, 10%). Results are shown for the clinical-only model trained on LUNA25, the clinical model augmented with MAISI-v2 synthetic CT (Clinical + MAISI-v2), and the clinical model augmented with NodMAISI synthetic CT (Clinical + NodMAISI). Note: "—" indicates model configurations that were not trained/evaluated at that training fraction.

| | Clinical training dataset percentage | | | |
|---|---|---|---|---|
| Fold | 100% | 50% | 20% | 10% |
| | Clinical (LUNA25) | | | |
| Fold 1 | 0.709 (0.673–0.742) | 0.684 (0.649–0.719) | 0.656 (0.618–0.693) | 0.656 (0.620–0.693) |
| Fold 2 | — | 0.658 (0.623–0.693) | 0.576 (0.541–0.611) | 0.511 (0.475–0.548) |
| Fold 3 | — | — | 0.609 (0.573–0.645) | 0.480 (0.443–0.517) |
| Fold 4 | — | — | 0.577 (0.539–0.614) | 0.519 (0.481–0.554) |
| Fold 5 | — | — | 0.584 (0.547–0.620) | 0.507 (0.472–0.544) |
| Fold 6 | — | — | — | 0.508 (0.476–0.544) |
| Fold 7 | — | — | — | 0.510 (0.475–0.546) |
| Fold 8 | — | — | — | 0.564 (0.526–0.601) |
| Fold 9 | — | — | — | 0.488 (0.453–0.526) |
| Fold 10 | — | — | — | 0.552 (0.516–0.588) |
| Mean | 0.708 | 0.671 | 0.600 | 0.529 |
| | Clinical + MAISI-v2 | | | |
| Fold 1 | 0.704 (0.670–0.738) | 0.678 (0.639–0.713) | 0.671 (0.634–0.708) | 0.645 (0.611–0.682) |
| Fold 2 | — | 0.639 (0.603–0.675) | 0.636 (0.600–0.672) | 0.658 (0.623–0.693) |
| Fold 3 | — | — | 0.624 (0.588–0.658) | 0.615 (0.579–0.653) |
| Fold 4 | — | — | 0.619 (0.582–0.655) | 0.642 (0.604–0.679) |
| Fold 5 | — | — | 0.622 (0.585–0.659) | 0.602 (0.566–0.636) |
| Fold 6 | — | — | — | 0.607 (0.571–0.642) |
| Fold 7 | — | — | — | 0.636 (0.604–0.671) |
| Fold 8 | — | — | — | 0.645 (0.606–0.679) |
| Fold 9 | — | — | — | 0.622 (0.582–0.658) |
| Fold 10 | — | — | — | 0.623 (0.587–0.659) |
| Mean | 0.704 | 0.659 | 0.634 | 0.630 |
| | Clinical + NodMAISI | | | |
| Fold 1 | 0.716 (0.681–0.750) | 0.687 (0.650–0.722) | 0.699 (0.665–0.733) | 0.671 (0.635–0.708) |
| Fold 2 | — | 0.678 (0.642–0.711) | 0.681 (0.641–0.719) | 0.646 (0.607–0.685) |
| Fold 3 | — | — | 0.639 (0.601–0.674) | 0.634 (0.598–0.669) |
| Fold 4 | — | — | 0.637 (0.601–0.673) | 0.665 (0.627–0.700) |
| Fold 5 | — | — | 0.671 (0.633–0.708) | 0.652 (0.612–0.691) |
| Fold 6 | — | — | — | 0.609 (0.573–0.644) |
| Fold 7 | — | — | — | 0.663 (0.625–0.700) |
| Fold 8 | — | — | — | 0.655 (0.617–0.690) |
| Fold 9 | — | — | — | 0.667 (0.628–0.701) |
| Fold 10 | — | — | — | 0.655 (0.618–0.691) |
| **Mean** | **0.716** | **0.682** | **0.666** | **0.652** |